\documentclass{article}

\usepackage{microtype}
\usepackage{graphicx}
\usepackage{booktabs} % for professional tables
\usepackage{multirow}
\usepackage{caption}
\usepackage{subcaption}

\usepackage{hyperref}
\usepackage{listings}
\usepackage[accepted]{icml2023}

\usepackage{amsmath}
\usepackage{amssymb}
\usepackage{mathtools}
\usepackage{amsthm}
\usepackage{makecell}

\usepackage[capitalize,noabbrev]{cleveref}

\theoremstyle{plain}

\theoremstyle{definition}

\theoremstyle{remark}

\usepackage[textsize=tiny]{todonotes}

\icmltitlerunning{Improving Visual Prompt Tuning for Self-supervised Vision Transformers}
\begin{document}

\twocolumn[
\icmltitle{Improving Visual Prompt Tuning for Self-supervised Vision Transformers}

% It is OKAY to include author information, even for blind
% submissions: the style file will automatically remove it for you
% unless you've provided the [accepted] option to the icml2023
% package.

% List of affiliations: The first argument should be a (short)
% identifier you will use later to specify author affiliations
% Academic affiliations should list Department, University, City, Region, Country
% Industry affiliations should list Company, City, Region, Country

% You can specify symbols, otherwise they are numbered in order.
% Ideally, you should not use this facility. Affiliations will be numbered
% in order of appearance and this is the preferred way.
\icmlsetsymbol{equal}{*}

\begin{icmlauthorlist}
\icmlauthor{Seungryong Yoo}{snuece}
\icmlauthor{Eunji Kim}{snuece}
\icmlauthor{Dahuin Jung}{snuece}
\icmlauthor{Jungbeom Lee}{snuece}
\icmlauthor{Sungroh Yoon}{snuece,snuai}
\end{icmlauthorlist}

\icmlaffiliation{snuece}{Electrical and Computer Engineering,}
\icmlaffiliation{snuai}{Interdisciplinary Program in Artificial Intelligence, Seoul National University, Seoul, Korea}

\icmlcorrespondingauthor{Sungroh Yoon}{sryoon@snu.ac.kr}
% \icmlcorrespondingauthor{Firstname2 Lastname2}{first2.last2@www.uk}

% You may provide any keywords that you
% find helpful for describing your paper; these are used to populate
% the "keywords" metadata in the PDF but will not be shown in the document
\icmlkeywords{Machine Learning, ICML}

\vskip 0.3in
]

% this must go after the closing bracket ] following \twocolumn[ ...

% This command actually creates the footnote in the first column
% listing the affiliations and the copyright notice.
% The command takes one argument, which is text to display at the start of the footnote.
% The \icmlEqualContribution command is standard text for equal contribution.
% Remove it (just {}) if you do not need this facility.

%\printAffiliationsAndNotice{}  % leave blank if no need to mention equal contribution
\printAffiliationsAndNotice{} % otherwise use the standard text.

\renewcommand\UrlFont{\rmfamily}

\begin{abstract}
Visual Prompt Tuning (VPT) is an effective tuning method for adapting pretrained Vision Transformers (ViTs) to downstream tasks. It leverages extra learnable tokens, known as prompts, which steer the frozen pretrained ViTs. Although VPT has demonstrated its applicability with supervised vision transformers, it often underperforms with self-supervised ones. Through empirical observations, we deduce that the effectiveness of VPT hinges largely on the ViT blocks with which the prompt tokens interact. Specifically, VPT shows improved performance on image classification tasks for MAE and MoCo v3 when the prompt tokens are inserted into later blocks rather than the first block. These observations suggest that there exists an optimal location of blocks for the insertion of prompt tokens. Unfortunately, identifying the optimal blocks for prompts within each self-supervised ViT for diverse future scenarios is a costly process. To mitigate this problem, we propose a simple yet effective method that learns a gate for each ViT block to adjust its intervention into the prompt tokens. With our method, prompt tokens are selectively influenced by blocks that require steering for task adaptation. Our method outperforms VPT variants in FGVC and VTAB image classification and ADE20K semantic segmentation. The code is available at \url{https://github.com/ryongithub/GatedPromptTuning}.
\end{abstract}

\section{Introduction}\label{sec:intro}
Currently, self-supervised learning (SSL) with Vision Transformers (ViTs)~\cite{bao2021beit, he2022masked, chen2021empirical, caron2021emerging} have exhibited remarkable results across diverse visual recognition tasks such as classification and semantic segmentation. SSL approaches strive to train the neural networks without biasing them toward labels containing very specific information about the corresponding tasks~\cite{zhao2020makes}. As a result, self-supervised models demonstrate superior scalability across various vision tasks compared to supervised ones~\cite{chen2020simple, he2020momentum, grill2020bootstrap, he2022masked, caron2021emerging, bao2021beit}. However, the efficacy of self-supervised models hinges on the chosen transfer learning strategy. For instance, a large performance gap exists between full fine-tuning and linear probing when employed as a transfer method for the Masked Autoencoder (MAE)~\cite{he2022masked}. The performance of SSL ViTs can vary significantly based on the transfer method, underscoring the importance of research into transfer strategies for SSL ViTs.

\begin{figure}[t]
\centering
\includegraphics[width=1.0
\linewidth]{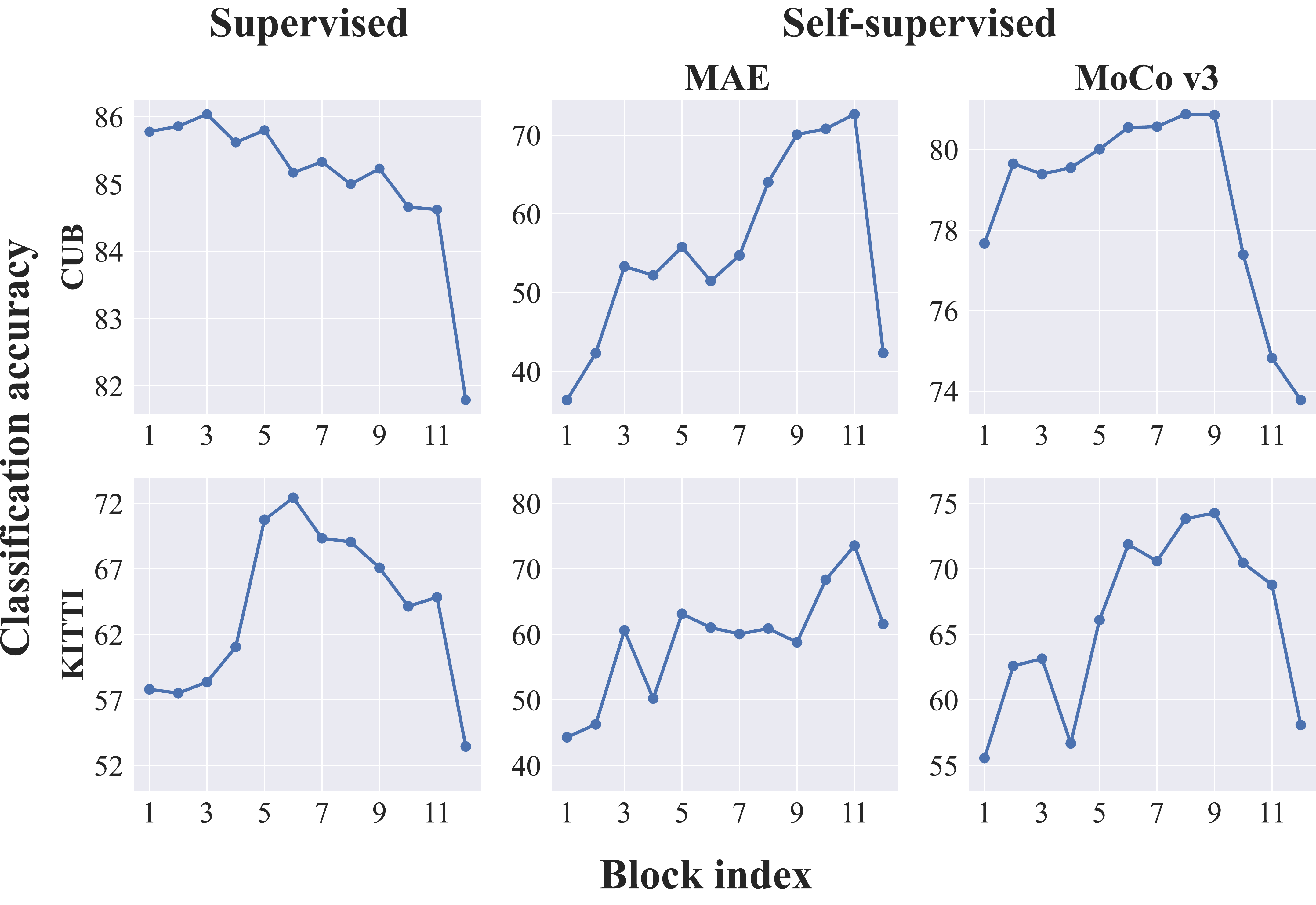}
\vspace*{-0.5cm}
\caption{Classification accuracy on the CUB and KITTI datasets with a varying location where prompt tokens are inserted in the pretrained ViT-B/16. MAE and MoCo v3 significantly improve their performances when prompt tokens are affected by the blocks after the 11\textsuperscript{th} and 8\textsuperscript{th} blocks, respectively. The block index denotes the initial insertion point of the prompt tokens.}
\label{fig:motivation_acc}
\end{figure}

Visual Prompt Tuning (VPT)~\cite{jia2022visual} is an effective prompt-based transfer learning technique, drawing its concept from previous work in the field of natural language processing (NLP)~\cite{li2021prefix, lester2021power}. In detail, VPT prepends learnable prompt tokens to the input sequences, which then act as task-specific instructions by steering the information from the fixed pretrained encoder. VPT, when used with supervised ViT backbones, has shown outstanding performance on numerous downstream tasks. However, despite the successes of prompt tuning~\cite{lester2021power, hambardzumyan2021warp, qin2021learning, huang2023extensibleacl} with SSL models in NLP~\cite{devlin2018bert, liu2019roberta}, VPT has demonstrated relatively poor performances with SSL ViTs~\cite{jia2022visual}. To address this, we propose a simple yet effective transfer method based on prompt tuning that enhances the performance of SSL ViTs.

It is well-known that Deep Neural Networks (DNNs) progressively abstract the information contained in data samples across their layers~\cite{lecun2015deep, schmidhuber2015deep, kriegeskorte2015deep}. Considering this in the context of VPT, information from all levels of abstraction influences the prompt tokens. However, these prompt tokens should be able to encode instructions by focusing solely on task-relevant information for target task adaptation. This can pose a challenge for prompt tokens from two perspectives. Firstly, the encoder possesses different information hierarchies across the blocks depending on how it was pretrained. Secondly, the task-relevant information required can vary depending on the downstream task. 

We conjecture that what the prompt tokens learn heavily relies on which blocks influence them during training. To justify our conjecture, we conducted an experiment to observe how performance varies depending on which blocks intervene with the prompt tokens. Interestingly, in Figure~\ref{fig:motivation_acc}, on the CUB~\cite{wah2011caltech} classification benchmark, MAE~\cite{he2022masked} and MoCo v3~\cite{chen2021empirical} boost their accuracy when the prompt tokens begin interacting with the 11\textsuperscript{th} and 8\textsuperscript{th} blocks in ViT-B, respectively. Particularly for MAE, the performance gap is as large as 36.4\%. Based on these findings, the key intuition of this study is that there exist desirable blocks that the prompt tokens should focus on to steer.

Depending on the pretraining strategy, the pretrained neural networks encode information differently in terms of the amount and content~\cite{zhao2020makes, bordes2021high}. Additionally, the relevant information required to solve the downstream task varies according to the task at hand. For these reasons, the task-relevant blocks will vary depending on their use in a downstream task and the pretraining method. Unfortunately, investigating all the possible cases to find the desirable sets of ViT blocks incurs substantial costs. To address this, we propose a simple yet effective method for learning to guide the prompt to selectively interact with desirable blocks that encode task-relevant information. We achieve this by introducing learnable gates for ViT blocks that adjust the intervention to the prompt tokens from  ViT blocks. With our proposed method, the prompt tokens focus on the blocks that need to be steered for the target task adaptation. 

Experimental results validate that our proposed method unlocks the potential of prompt tuning as a universal tuning strategy for self-supervised ViTs. On the FGVC benchmark, which encompasses five fine-grained classification tasks~\cite{wah2011caltech, van2015building, khosla2011novel, gebru2017fine, nilsback2008automated}, we achieve an average accuracy of 73.4\% for MAE and 83.0\% for MoCo v3. These results substantially outperform VPT-shallow and are either outperforming or comparable to VPT-deep. Moreover, on the VTAB-1K  benchmark~\cite{zhai2019large}, which consists of 19 diverse visual classification tasks, our proposed method attains an average accuracy of 49.2\% and 65.8\%  for MAE and MoCo v3, respectively, significantly outperforming both VPT-deep and VPT-shallow. Our method demonstrates its strength not only in classification tasks but also in dense prediction tasks such as semantic segmentation. It exhibits superior performance compared to VPT counterparts on the ADE 20K semantic segmentation benchmark~\cite{zhou2017scene}, achieving 38.4 mIoU for MAE and 36.8 mIoU for MoCo v3.
\section{Preliminaries}
\textbf{Vision Transformer}. 
Typically, ViT consists of a patch embedding layer, a stack of $L$ transformer blocks, and a classification head. To process an image of a height $H$, a width $W$, and a channel $C$ into a ViT, we divide the image as a grid of patches $x_i\in \mathbb{R}^{P\times P\times C}$, where $P$ is a patch size and $i=1, \dots, \frac{HW}{P^2}$. Each $x_i$ is embedded as a $D$-dimensional feature and $D$-dimensional positional embedding is added to each patch token to provide position information to ViTs as follows:
\begin{equation}
\begin{gathered}\label{eq:ViT}
    z_i^0 = \text{PatchEmbed}(x_i) + e_i,   \\
\end{gathered}
\end{equation}
where $z_i^0$ denotes the embedded input token for the first ViT block and $e_i$ denotes the positional embedding. Let the input patch tokens for $l$\textsuperscript{th} block as $\textbf{Z}^{l-1}=\big[z_1^{l-1}, \dots, z_N^{l-1}\big]$, where $N=\frac{HW}{P^2}$, $l=1, \dots, L$, and $L$ is the number of blocks. Patch tokens and an additional learnable token for classification, $z_{\text{CLS}}^l$, is fed to the blocks as follows:
\begin{equation}
\begin{gathered}\label{eq:ViT-block}
    \big[z_{\text{CLS}}^l, \textbf{Z}^l\big] = \text{Block}^l\Big(\big[z_{\text{CLS}}^{l-1}, \textbf{Z}^{l-1}\big]\Big) \in \mathbb{R}^{(N+1) \times D},
\end{gathered}
\end{equation} 
where each block consists of a multi-head self-attention followed by a feed-forward layer with layer normalization~\cite{layernorm} and residual connection~\cite{resnet}. Among multiple self-attention heads in $l$\textsuperscript{th} block, a single self-attention head~\cite{vaswani2017attention} is formulated as follows:
% Self-attention
\begin{equation}
\begin{gathered} \label{eq:attn}
    \text{Attention}(Q^l, K^l, V^l) = a^l V^l \\
    \text{s.t. } a^l = \text{Softmax}\Big(Q^l{K^l}^T\Big),
\end{gathered}
\end{equation}
where $Q^l, K^l, V^l$ present the input query, key, and value tokens built by the linear projection of $\big[z_{\text{CLS}}^{l-1}, \textbf{Z}^{l-1}\big]$, respectively, and $a^l$ presents the self-attention score calculated in $l$\textsuperscript{th} block. 
Lastly, the classification head has a single feed-forward layer for class prediction.

% VPT-shallow
\textbf{Visual Prompt Tuning}. 
VPT~\cite{jia2022visual} trains continuous prompts in the embedded space. Learnable prompt tokens $\textbf{P}=[p_1, \dots, p_{N_p}] \in\mathbb{R}^{N_P \times D}$ are prepended to the input sequence, where $N_p$ is the number of prompt tokens and $D$ is the dimension of the prompt token. During transfer learning, the prompt tokens and a classiﬁcation head are only trainable and the pretrained ViT encoder is fixed. The prompt tokens learn to encode task-specific instructions by interacting with patch representations across all ViT blocks. VPT introduced two variants: VPT-shallow and VPT-deep. VPT-shallow inserts learnable prompt tokens as input in the first block. The following equations formulate the VPT-shallow:
\begin{equation}
\begin{gathered}\label{eq:VPT}
    \big[z_{\text{CLS}}^1, \textbf{Z}_P^1, \textbf{Z}^1\big] = \text{Block}^1\Big(\big[z_{\text{CLS}}^{0}, \textbf{P}, \textbf{Z}^{0}\big]\Big)\\
    \big[z_{\text{CLS}}^l, \textbf{Z}_P^l, \textbf{Z}^l\big] = \text{Block}^l\Big(\big[z_{\text{CLS}}^{l-1}, \textbf{Z}_P^{l-1}, \textbf{Z}^{l-1}\big]\Big),
\end{gathered}
\end{equation}
where $\textbf{Z}_P^l$ denotes the output prompt representation of $l$\textsuperscript{th} block.
Differently, VPT-deep injects block-wise learnable prompt tokens $\textbf{P}^{l-1}$ to each block, not the previous block's output $\textbf{Z}_P^{l-1}$:
\begin{equation}
\begin{gathered}\label{eq:VPT-deep}
    \big[z_{\text{CLS}}^l,\textbf{\underline{\hspace{0.25cm}}} , \textbf{Z}^l\big] = \text{Block}^l\Big(\big[z_{\text{CLS}}^{l-1}, \textbf{P}^{l-1}, \textbf{Z}^{l-1}\big]\Big).
\end{gathered}
\end{equation}

\section{Motivation}\label{sec:motivation}
The pretrained knowledge and the downstream task are two important factors in transfer learning scenarios. In this section, we discuss the motivational background of our research from these two perspectives.

\begin{figure}[t]
\centering
\includegraphics[width=1.\linewidth]{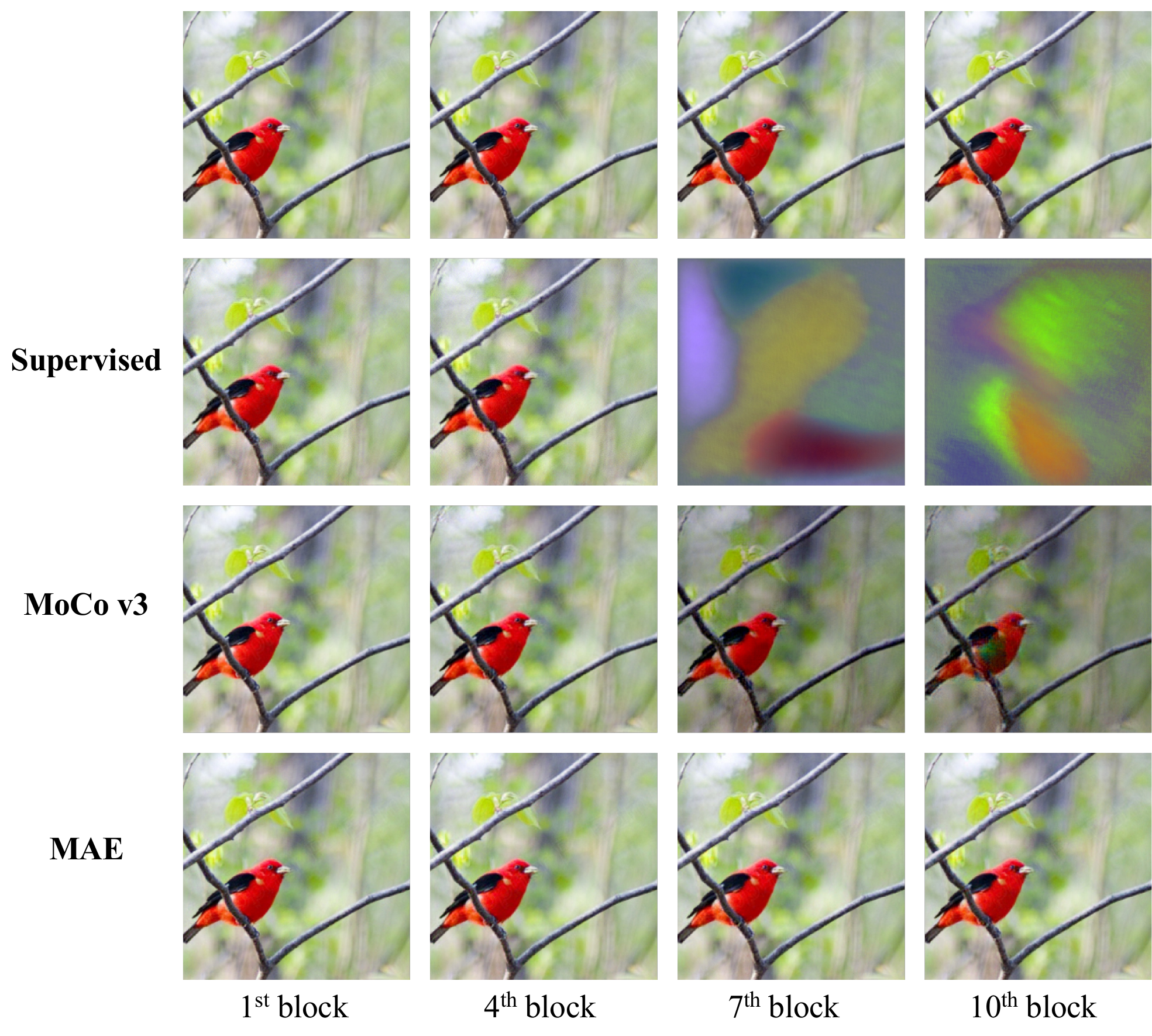}
\vspace*{-0.6cm}
\caption{Reconstructed images using Deep Image Prior (DIP) with pretrained ViT block's representation as a training target. The reconstructed image maintains its similarity to the original image as the block preserves information till the last block. \textbf{Row 1}: original image. \textbf{Rows 2-4}: reconstruction results for each pretrained ViTs.  Poor results in late blocks (7\textsuperscript{th} and 10\textsuperscript{th}) of the supervised model indicate that it discards more information across blocks than the self-supervised ViTs.}
\label{fig:dip}
\vspace*{-0.4cm}
\end{figure}

\textbf{Pretrained knowledge}.
We find that the block where the prompt tokens are inserted at first leads to varying performances in Figure~\ref{fig:motivation_acc}. We verify these findings from the perspective of information contained in each block of ViTs. We utilize Deep Image Prior (DIP)~\cite{ulyanov2018deep} to investigate the information change across the blocks of the pretrained ViTs. DIP reconstructs an image by updating the random noise so that the original image's representation and the updated noise's representation are close in the representation space. Using the reconstructed image, we can infer the information that the representation space of each neural network unit encodes. As shown in Figure~\ref{fig:dip}, in the later blocks, supervised ViT tends to discard more information than self-supervised ViT. In contrast to supervised one, self-supervised ViTs, MoCo v3~\cite{chen2021empirical} and MAE~\cite{he2022masked} retain rich information across the blocks. This tendency is also evident in the reconstructed image quality scores, such as PSNR and SSIM, as indicated in Figure~\ref{fig:ps_score} in Appendix~\ref{subsec:emp_quan}. Furthermore, it can be observed that even self-supervised Vision Transformers exhibit differences in the information content encoded by each block. When comparing MoCo v3 to MAE, it is noticeable that MoCo v3 exhibits a decrease in color information after the middle block. More DIP results are available in Appendix~\ref{subsec:dip_results}.

This example shows that different pretraining methods lead ViT blocks to differ in terms of the amount and content of information they encode. It follows that the locations of blocks containing task-relevant information vary depending on the pretrained ViTs. If the prompt is inserted into the first block without considering these differences, the accumulated intervention of the task-irrelevant blocks could disturb the prompt tokens to focus on the task-relevant blocks.

\textbf{Task diversity}.
Task-relevant information varies depending on the downstream task. For example, in the case of classification on the CUB dataset, discriminating fine color and shape information is crucial to classify diverse bird species. Unlike the bird classification task with CUB, the KITTI~\cite{geiger2013vision} distance task requires capturing position and scale information to accurately estimate the distance to the objects in the scene. Thus,  even with the same pretrained Vision Transformer, the locations of blocks encoding task-relevant information can vary depending on the task. Figure~\ref{fig:motivation_acc} illustrates that the performance change according to the location of the prompt insertion is different between CUB and KITTI. This indicates that the blocks that are desirable for the prompt insertion to perform task-adaptation are task-dependent.

\section{Proposed Method}\label{sec:method}
\begin{figure}[t]
\centering
\includegraphics[width=0.85
\linewidth]{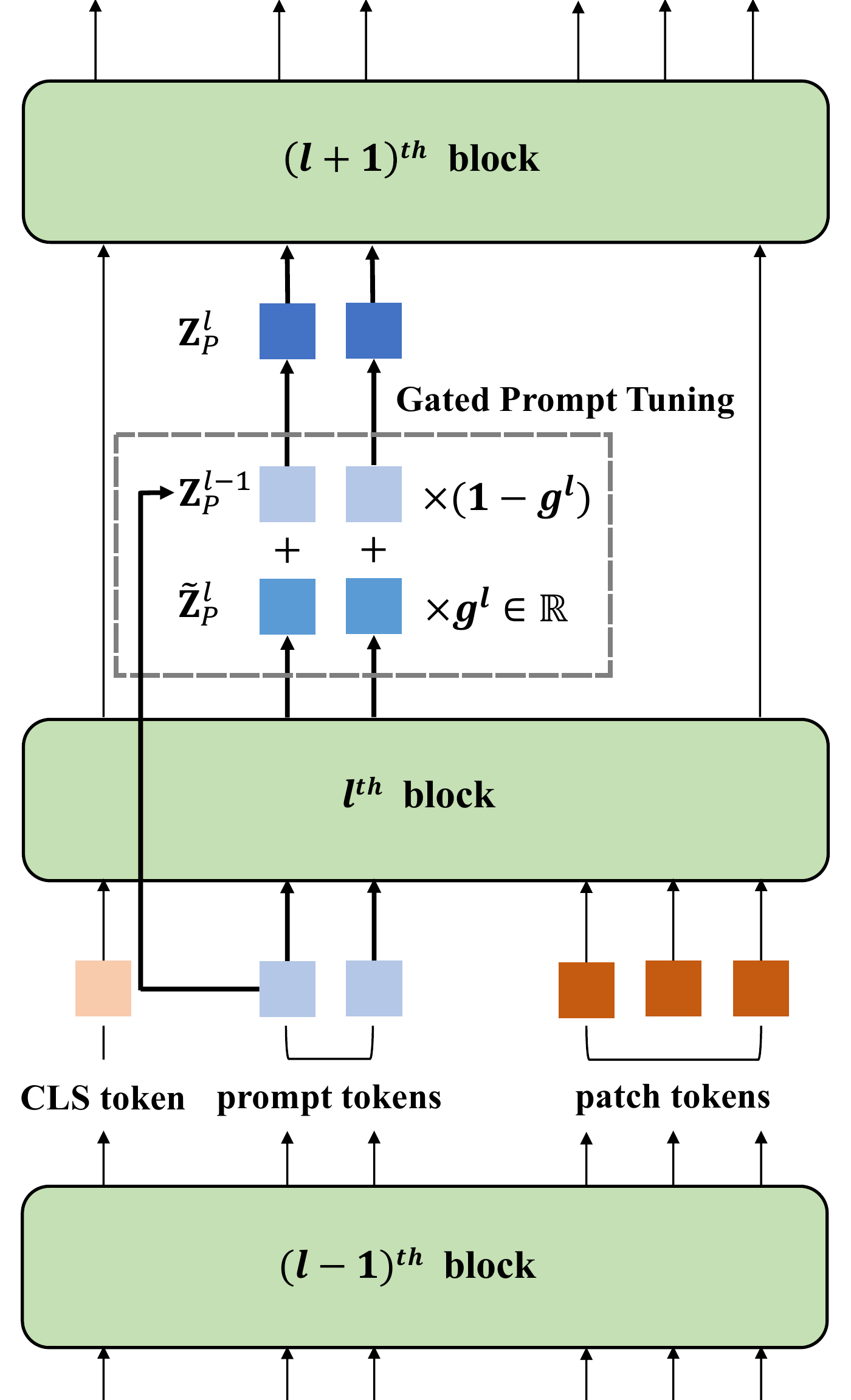}
\vspace*{-0.2cm}
\caption{An illustration of our proposed method, \textbf{Gated Prompt Tuning}. $\textbf{Z}_P^{l-1}$ and $\tilde{\textbf{Z}}_P^l$ are input and output prompt representations of $l$\textsuperscript{th} block. The learnable gate $g^l$ convexly combinates $\textbf{Z}_P^{l-1}$ and $\tilde{\textbf{Z}}_P^l$ so that the $(l+1)$\textsuperscript{th} block receives the prompt representation $\textbf{Z}_P^l$ in which the intervention of $l$\textsuperscript{th} block into the prompt representation is adjusted by $g^l$.}
\label{fig:model}
\vspace*{-0.5cm}
\end{figure}

In the previous section, we discuss that the locations of the blocks where the prompt can derive improved performances depend on the SSL method. Further, since the required task-specific instruction may vary depending on the task, the desirable blocks for the prompt could be different in the same pretrained model.
Sufficient task performance may not be secured when prompts are inserted from the first block without careful consideration of this difference. In this respect, we consider a universal prompt tuning method that can learn to select the blocks where steering for task adaptation is strongly required. To achieve this goal, we introduce Gated Prompt Tuning, which leverages learnable gates to adjust each block's intervention into the prompt tokens. The learnable gates enable prompts to readily focus on the blocks that require steering for task adaptation. Similar to VPT-shallow, our method prepends the prompt tokens only once to the input patch sequence. Our framework is illustrated in Figure~\ref{fig:model}.

\subsection{Gated Prompt Tuning}
First, we define gate priors $\Gamma = [\gamma^1,\dots, \gamma^{L-1}]$, a set of scalar values for all blocks except for the last block. After being scaled with a sigmoid function, the gate prior is utilized as a gate value of $l$\textsuperscript{th} block:
\begin{equation}
\begin{gathered}\label{eq:gamma}
    g^l = {1\over{1 + e^{-\gamma^l}}} \in \mathbb{R
}.\\
\end{gathered}
\end{equation}
The gating operation for the input prompt representations at $(l+1)$\textsuperscript{th} block (\textit{i.e.,} $\textbf{Z}_P^l$) is formulated as follows:
\begin{equation}
\begin{gathered}\label{eq:gate}
    \big[z_{CLS}^l, \tilde{\textbf{Z}}_P^l, \textbf{Z}^l\big] = \text{Block}^l\Big(\big[z_{CLS}^{l-1}, \textbf{Z}_P^{l-1}, \textbf{Z}^{l-1}\big]\Big),\\
    \textbf{Z}_P^l = g^l \cdot \tilde{\textbf{Z}}_P^l + (1 - g^l)\cdot \textbf{Z}_P^{l-1},
\end{gathered}
\end{equation}
where $\tilde{\textbf{Z}}_P^l$ denotes the output prompt representation of $l$\textsuperscript{th} block. In Eq.~\ref{eq:gate}, $(l+1)$\textsuperscript{th} block receives the weighted sum of $\textbf{Z}_P^{l-1}$ and $\tilde{\textbf{Z}}_P^l$, which are the input and output prompt representation of $l$\textsuperscript{th} block, respectively. Here, the gate value $g^l$ controls
the contribution on composing the input prompt representation of $(l+1)$\textsuperscript{th} block between $\textbf{Z}_P^{l-1}$ and $\tilde{\textbf{Z}}_P^l$. Therefore, gate $g^l$ determines how much of the previous block's influence on the prompt is carried over to the next block. During training, the gates learn to adjust the intervention of each block in the prompt tokens, which enables the prompts to focus on desirable blocks that require task-specific steering.

Right before the task head, the last block refines the representations to be discriminative for the task. The input prompt representations ($i.e.$ $\textbf{Z}_P^{L-1}$) for the last block have a significant role in this regard. 
Using Eq.~\ref{eq:gate}, we can express the input prompt representations of the last block as follows:
\begin{equation}
\begin{aligned}\label{eq:momentum}
    \textbf{Z}_P^{L-1} &= \bigg(\prod_{l=1}^{L-1} (1-g^l)\bigg)\textbf{P}\\
    &+\sum_{l=1}^{L-2} \bigg(\prod_{m=l+1}^{L-1} (1-g^m)\bigg)g^l  \tilde{\textbf{Z}}_P^l + g^{L-1} \tilde{\textbf{Z}}_P^{L-1}.
\end{aligned}
\end{equation}
Since $\tilde{\textbf{Z}}_P^l$ is the output prompt representations of $l$\textsuperscript{th} block, Eq.~\ref{eq:momentum} can be interpreted as a selective aggregation of all output prompt representations from the ViT blocks by the learned gates. The selective aggregation results in adaptive instructions for target tasks.

\subsection{Adaptive Attention Shaping}\label{subsec:lt}
Prompt tuning can be understood as learning to steer the behavior of ViTs by manipulating the pretrained attention score of the patch tokens by extending input sequences with extra prompt tokens. Based on this intuition, as an additional technique to boost the task-adaptability of prompt tuning, we introduce Adaptive Attention Shaping. We define learnable temperature $\mathcal{T} = [\tau^1, \dots, \tau^L]$, a set of scalar values that adjust the attention value in the self-attention operation of the corresponding block. With the learnable temperature, we rewrite the self-attention score in Eq.~\ref{eq:attn} as follows: 
\begin{equation}
\begin{gathered}\label{eq:temp_learn}
    a^l = \text{Softmax}\Big(\frac{Q^l{K^l}^T}{\tau^l} \Big).
\end{gathered}
\vspace{0.2cm}
\end{equation}
$\mathcal{T}$ directly reshape the blocks' self-attentions by making them sharper or smoother so that $\mathcal{T}$ aids prompts to encode beneficial instruction to solve the current task.

\begin{table*}[t]
\caption{Classification results on FGVC. TOTAL PARAMS denotes the total number of parameters for all tasks including the backbone encoder ViT-B, prompt tokens and the task heads. Bold fonts denote the best performance in each benchmark.} 
\label{table:fgvc}
\vskip 0.15in
\begin{center}
\begin{small}
\begin{sc}
\begin{tabular}{llccccccc}
\toprule
\multirow{2}{*}{SSL}               & \multirow{2}{*}{Method}  & \multicolumn{1}{p{1cm}}{\centering total \\ params} & \multirow{2}{*}{CUB}   & \multirow{2}{*}{Flowers} & \multirow{2}{*}{Cars}  & \multirow{2}{*}{Dogs}  & \multirow{2}{*}{NABirds} & \multirow{2}{*}{AVG} \\
\midrule
\multirow{3}{*}{MAE}  & VPT-Shallow     &1.02$\times$                     & 42.15          & 69.15             & 43.38             & 77.07          & 57.43            & 57.84 \\
& VPT-Deep     & 1.02$\times$                     & 68.33          & \textbf{80.05}             & 67.67             & 78.83          & 65.22            & 72.02 \\
                      & Ours    &1.02$\times$                      & \textbf{70.56}  &78.55   & \textbf{71.70}              & \textbf{78.9} & \textbf{67.26}            & \textbf{73.39} \\
                      \midrule
\multirow{3}{*}{MoCo v3} & VPT-Shallow    &1.02$\times$  &79.05          & 90.47           & 71.91            & 81.97      & 72.92           & 79.26 \\
& VPT-Deep     &1.02$\times$                      & 82.67          & \textbf{94.41}           &\textbf{79.18}            & 83.33      & 75.99           & \textbf{83.12} \\
                      & Ours    &1.02$\times$                     & \textbf{82.86} &93.71  &79.02 &\textbf{83.37}  & \textbf{76.02}          &83.00 \\

\bottomrule
\end{tabular}
\end{sc}
\end{small}
\end{center}
\vskip -0.1in
\end{table*}

\begin{table*}[t]\centering
\caption{Classification results on VTAB-1K. TOTAL PARAMS denotes the total parameters for all tasks, including the backbone encoder ViT-B, prompt tokens, and the task heads. ($\dag$) denotes the reported performances in the original paper of VPT~\cite{jia2022visual}. Bold fonts denote the best performance in each benchmark. } 
\label{table:vtab}
\vskip 0.15in
\begin{center}
\begin{small}
\begin{sc}
\begin{tabular}{llccccc}
\toprule
\multirow{2}{*}{SSL} &\multirow{2}{*}{Method} &\multicolumn{1}{p{1cm}}{\centering total \\ params} &\multirow{2}{*}{Natural (7)} &\multirow{2}{*}{Specialized (4)} &\multirow{2}{*}{Structured (8)} &\multirow{2}{*}{AVG}\\
\midrule
\multirow{3}{*}{MAE} &VPT-shallow &1.01$\times$ &39.96\rlap{$^\dag$} &69.65\rlap{$^\dag$} &27.50\rlap{$^\dag$} &40.96\rlap{$^\dag$}\\
&VPT-deep &1.04$\times$ &36.02\rlap{$^\dag$} &60.61\rlap{$^\dag$} &26.57\rlap{$^\dag$} &37.22\rlap{$^\dag$}\\
&Ours &1.01$\times$ &\textbf{47.61} &\textbf{76.86} &\textbf{36.80} &\textbf{49.22}\\
\midrule
\multirow{3}{*}{MoCo v3} &VPT-shallow &1.01$\times$ &67.34\rlap{$^\dag$} &82.26\rlap{$^\dag$} &37.55\rlap{$^\dag$} &57.94\rlap{$^\dag$}\\
&VPT-deep &1.01$\times$ &70.27\rlap{$^\dag$} &83.04\rlap{$^\dag$} &42.38\rlap{$^\dag$} &61.22\rlap{$^\dag$}\\
&Ours &1.01$\times$ &\textbf{74.84} &\textbf{83.38} &\textbf{49.10} &\textbf{65.80}\\
\bottomrule
\end{tabular}
\end{sc}
\end{small}
\end{center}
\end{table*}

\subsection{Comparison with VPT}

Our method incorporates learnable gates to regularize prompt tokens to interact with task-relevant blocks and enables effective prompt learning by utilizing the additional capacity obtained through learnable temperatures. However, in VPT, there is no such consideration for effective learning of the prompt tokens.

Another difference between our method and VPT lies in whether the prompt can provide sample-specific but task-relevant instruction. In VPT-shallow, the prompt representation passed to each block is conditioned on the patch representation from the previous block, allowing for sample-specific instructions at each block. VPT-shallow can be considered as a special case of our method, as all learned gates are set to 1, and it lacks a learnable temperature. However, when applied to self-supervised ViTs, the prompt tokens interact with all blocks in VPT-shallow, and thus, VPT-shallow has a limitation in effectively targeting task-relevant information.

VPT-deep addresses the differences in task relevancy across blocks by providing independent learnable prompt token sets for each block. The prompts in each block of VPT-deep are trained to provide task-relevant instructions on average over the entire training data. However, since all samples receive the same instruction from shared prompt tokens at each block, VPT-deep could not provide sample-specific instruction.

In our method, the gating operation allows the prompt token to selectively interact with the blocks, and the input prompt representation for each block is dependent on the patch representation from the preceding block. Therefore, our method enables providing sample-specific but task-relevant instructions at each block.

\section{Experiments}\label{sec:experiments}
\subsection{Experimental Setup}
\textbf{Downstream tasks and datasets}.
We evaluate our method with two types of downstream tasks: image classification and semantic segmentation. For image classification, we conduct experiments on FGVC and VTAB-1K~\cite{zhai2019large} benchmark. FGVC includes five fine-grained classification tasks: CUB~\cite{wah2011caltech}, Oxford Flowers~\cite{nilsback2008automated}, Stanford Cars~\cite{gebru2017fine}, Stanford Dogs~\cite{khosla2011novel} and NABirds~\cite{van2015building}. VTAB-1K is divided into three subgroups: \textit{Natural} with natural images, \textit{Specialized} with images obtained from specialized equipment, and \textit{Structured} which requires structural understanding such as 3D depth prediction. 

For semantic segmentation, we evaluate the performances on ADE20K~\cite{zhou2017scene} benchmark which contains 20K images with 150 object categories. For the segmentation model, we use SETR-PUP~\cite{zheng2021rethinking} which utilizes ViT~\cite{dosovitskiy2020image} as a backbone encoder. Note that the original implementation of SETR-PUP adopts four auxiliary heads at the 10\textsuperscript{th}, 15\textsuperscript{th}, 20\textsuperscript{th}, and 24\textsuperscript{th} blocks of ViT-L. Since we use ViT-B/16 in all experiments, two auxiliary heads are used at the 5\textsuperscript{th} and 9\textsuperscript{th} blocks. Further experimental details are described in Sec.~\ref{detail_hyp} in Appendix.

\textbf{Self-supervised Vision Transformers}.
Our study utilizes two well-performing self-supervised Vision Transformers, MAE~\cite{he2022masked} and MoCo v3~\cite{chen2021empirical}, which are pretrained on ImageNet-1K~\cite{deng2009imagenet}. Pretrained model parameters are taken from the official repository of Visual Prompt Tuning~\cite{jia2022visual}. We use ViT-B/16 as the backbone architecture in all experiments of this study.

\subsection{Main Results}
\subsubsection{Classification on FGVC}
We evaluate the fine-grained classification performance in the FGVC benchmark. For VPT-shallow and our method, we set 100 tokens, and for VPT-deep, we set 10 tokens for each block, which means that 120 tokens are totally used for VPT-deep. Table~\ref{table:fgvc} shows that our method consistently outperforms the VPT-shallow counterpart by a large margin both for MAE and MoCo v3 in all datasets. This shows that our approach leads the prompt to encode enhanced instruction in SSL ViTs for task adaptation. Note that compared to VPT-deep, MAE with our method has higher average accuracy while surpassing in almost all datasets. When applied to MoCo v3, even though VPT-deep uses 20\% more extra parameters, our method shows comparable performance to VPT-deep on average. According to this, when the number of added prompt tokens is similar, our method is more effective at utilizing prompt tokens for fine-grained classification than both of these two variants of VPT. Moreover, our method shows more efficient results with fewer prompt tokens compared to VPT-deep. The additional experimental results for FGVC with fewer tokens can be found in Table~\ref{table:fgvc-token} in Appendix~\ref{subsec:token_eff}.

\subsubsection{Classification on VTAB-1K}
With the VTAB-1K benchmark, we test our method’s capability of driving the backbone encoder to capture generic visual concepts on three distinct groups of datasets. Table~\ref{table:vtab} shows the classification results on VTAB-1K. MAE with our approach largely outperforms the VPT-deep and VPT-shallow in all three groups. Also, our method provides substantial gains for MoCo v3 in \textit{Natural} and \textit{Structured} groups compared to VPT counterparts. Especially for both MAE and MoCo v3, we observe the largest performance boosts in \textit{Structured} group, 10.23\% and 6.72\% from VPT-deep, respectively. This indicates that our method learns prompt tokens that enable SSL ViTs to transfer more effectively. It is effective not only with natural images but also in situations where image domains are different or when structural comprehension is required. 

\begin{table}[t]\centering
\setlength{\tabcolsep}{4pt}
\caption{Semantic segmentation results on ADE20K with SETR-PUP~\cite{zheng2021rethinking} as the segmentation model. For VPT-deep, $(\times 12)$ denotes that the same number of prompt tokens are used for each block of ViT-B/16. PT denotes prompt token. Bold fonts denote the best performance in each metric.}
\label{table:ade}
\vskip 0.15in
\begin{center}
\begin{small}
\begin{sc}
\begin{tabular}{lllcc}
\toprule
\multirow{2}{*}{SSL} &\multirow{2}{*}{Method} & \multirow{2}{*}{\# of PTs} &\multicolumn{1}{p{1cm}}{\centering mIou \\ (SS)} &\multicolumn{1}{p{1cm}}{\centering mIoU \\ (ms)} \\
\midrule
\multirow{3}{*}{MAE} &VPT-shallow &100 &34.20 &35.23 \\
&VPT-deep &10\rlap{$(\times 12)$}  &37.76 &38.80 \\
&Ours  &100 &\textbf{38.44} &\textbf{39.81} \\
\midrule
\multirow{3}{*}{MoCo v3} &VPT-shallow &100 &34.55 &36.18 \\
&VPT-deep &10\rlap{$(\times 12)$}  &35.50 &37.15 \\
&Ours  &100 &\textbf{36.81} &\textbf{38.55} \\
\bottomrule
\end{tabular}
\vspace{-0.3cm}
\end{sc}
\end{small}
\end{center}
\end{table}
\subsubsection{Semantic segmentation on ADE20K}
To validate that the applicability of our method is not limited to image classification, we evaluate the semantic segmentation task on ADE20K. Following the settings in VPT~\cite{jia2022visual}, we used SETR-PUP~\cite{zheng2021rethinking} for the segmentation model. As shown in Table~\ref{table:ade}, our method brings large performance gains from VPT-shallow for both MAE and MoCo v3. When compared to VPT-deep, MAE and MoCo v3 with our method also shows advanced results. Note that in this experiment, our method beats all VPT counterparts while VPT-deep uses 20\% more prompt tokens. This implies that our method utilizes prompt tokens more effectively in semantic segmentation tasks. In Appendix~\ref{subsec:token_eff}, we present the results for ADE20K when using fewer prompt tokens in Table~\ref{table:ade-token}. In summary, as shown in the FGVC, VTAB-1K, and ADE20K benchmark results, our proposed method is a prompt-based transfer strategy that better exploits SSL ViTs for diverse vision tasks.

\subsection{Additional Analysis}
\textbf{Analysis on the learned gates}.
As we explained in Eq.~\ref{eq:momentum}, our method is interpreted as a selective aggregation of the prompt representation from each block. Using the learned gate values, $g^l$, we are able to determine the contribution of each block to the prompt delivered to the last block. In Eq.~\ref{eq:momentum}, the weight applied to the output prompt representation of each block is expressed as follows:
\begin{equation}
\begin{gathered}\label{eq:accumul_g}
    \tilde{g}^l = \Big(\prod_{m=l+1}^{L-1}(1-g^m)\Big)g^l,\ l=1, \dots, L-2 \\
    \tilde{g}^{L-1} = g^{L-1}
\end{gathered}
\end{equation}
Based on the weight $\tilde{g}^l$, we define the selection ratio, which represents the influence of each block on the prompt representation of the last block:
\begin{equation}
\begin{gathered}\label{eq:selection}
    \mathbf{r}^l = {\tilde{g}^l \over{ \sum_{m=1}^{L-1}\tilde{g}^m}}.
\end{gathered}
\end{equation}

\begin{figure}[t]
\centering
\includegraphics[width=1.0
\linewidth]{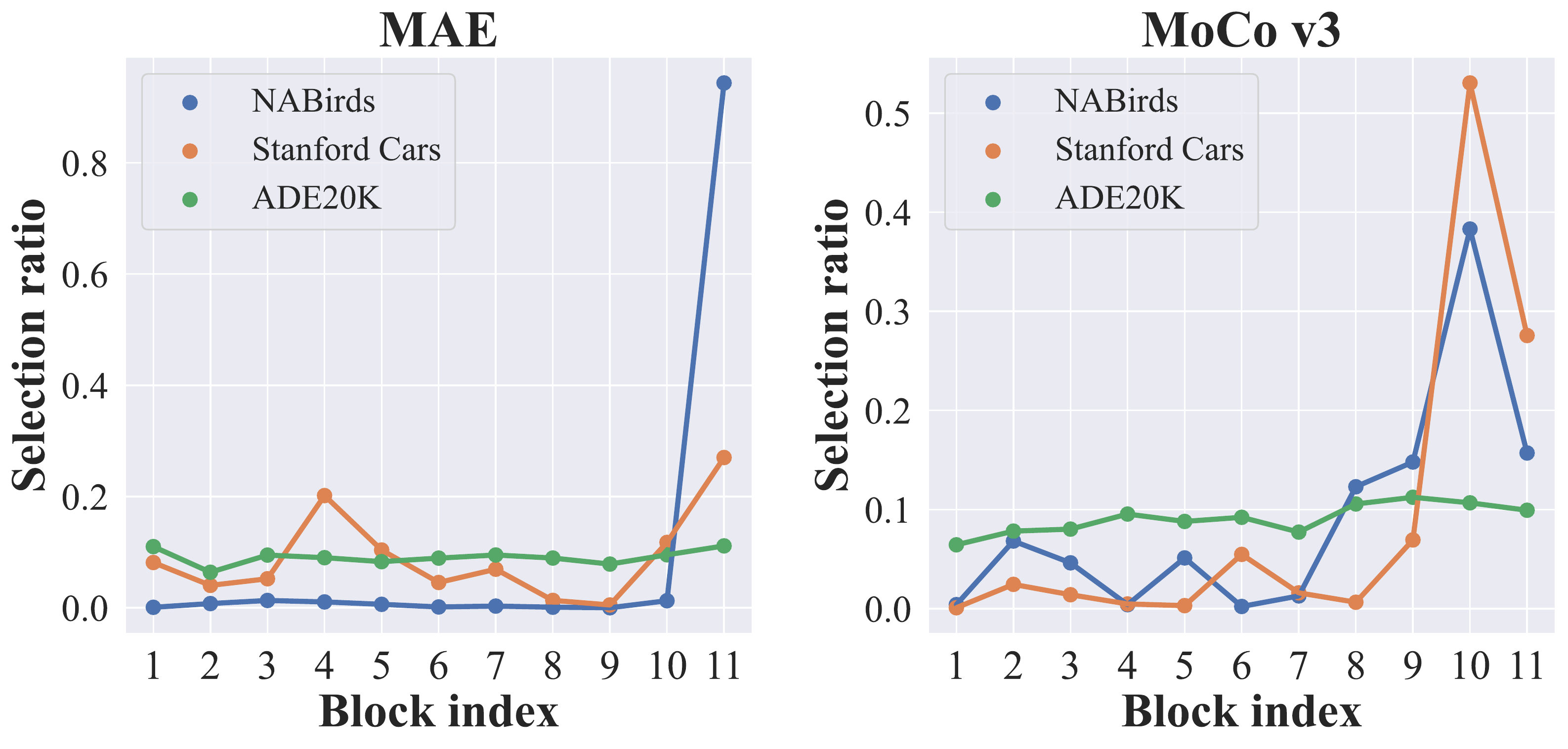}
\vspace*{-.6cm}
 \caption{Selection ratio $\mathbf{r}$ on the NABirds, Stanford Cars fine-grained classification and ADE20K semantic segmentation. The selection ratio represents the influence of each block on the prompt representation of the last block.}
  \label{fig:selection}
\end{figure}

Figure~\ref{fig:selection} shows the selection ratio calculated from models trained for different transfer scenarios. As shown in the figure, the selection ratio varies according to the downstream task and SSL method. For instance, the prompts learned for MAE focus primarily on the 11\textsuperscript{th} block for the NABirds dataset while focusing almost equally on the 4\textsuperscript{th} and 11\textsuperscript{th} blocks for Stanford Cars dataset. In addition, we observe that the learned prompts focus on different blocks depending on the SSL ViT. On NABirds and Stanford Cars, learned prompts with MAE tend to focus on the 11\textsuperscript{th} block, while with MoCo v3, they tend to focus on the 10\textsuperscript{th} block. In other words, the gates guide the prompts to take into account the information change across the blocks that differs depending on the self-supervised method. In ADE20K segmentation, the learned prompts appear to have uniform ratios across the blocks. This indicates that the gates learn to make prompts to steer all the blocks since multi-level information is advantageous for segmentation tasks~\cite{lin2017feature}.
These differences in selection ratios support our motivation that prompts should focus on different blocks so that they can be selectively affected by ViT blocks for effective task adaptation.

\begin{figure}[t]
\centering
\includegraphics[width=1.0
\linewidth]{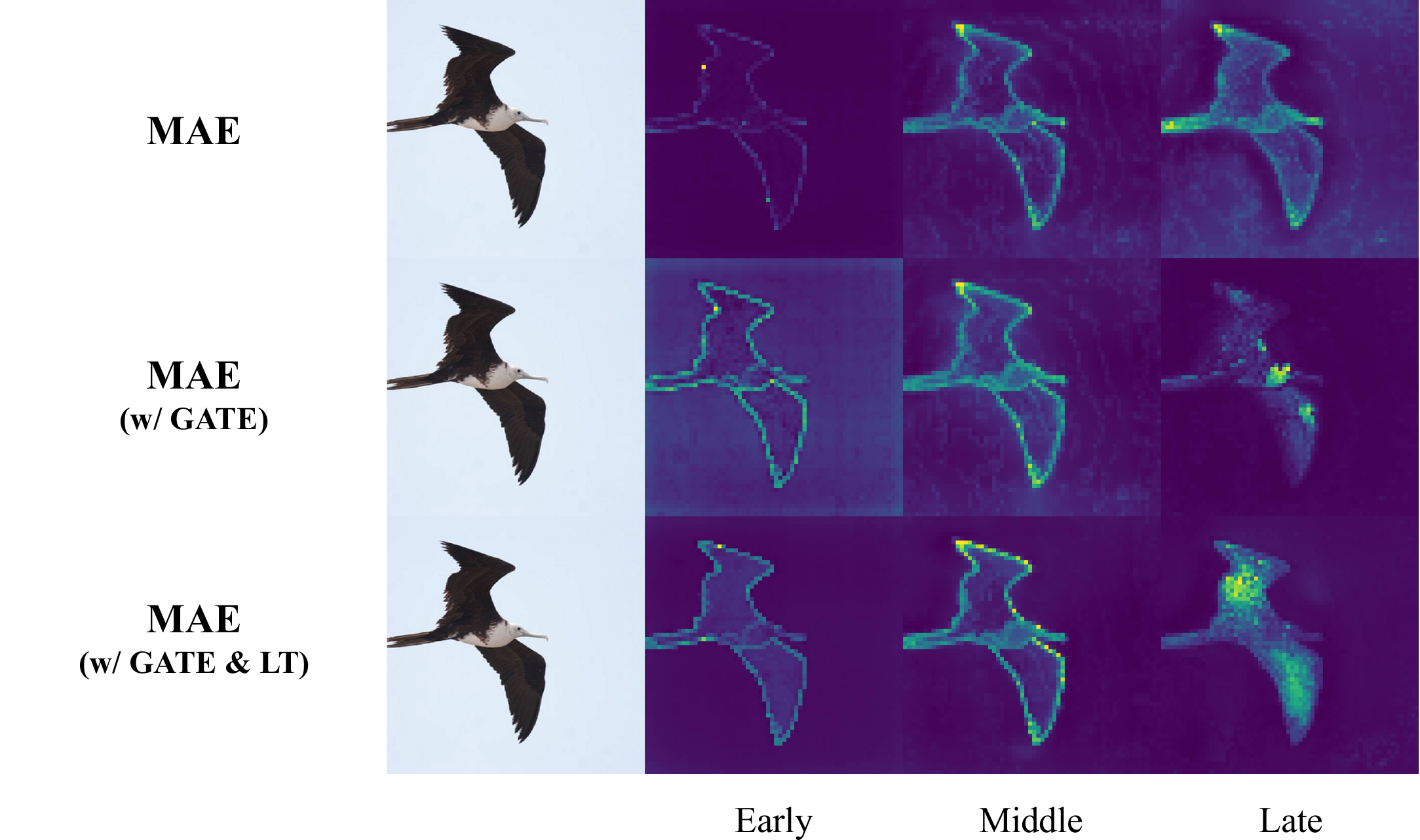}
\vspace*{-.5cm}
\caption{Visualization on self-attention map of ViT-B/16 blocks. Both prompt tuning and temperature scaling adjust the self-attention map from MAE. GATE denotes Gated Prompt Tuning and LT denotes Adaptive Attention Shaping with learnable temperatures.}
\label{fig:gate_lt}
\vspace*{-0.2cm}
\end{figure}

\textbf{Adjusted Self-attention}.
 As we discussed in Section~\ref{subsec:lt}, prompt tuning manipulates self-attention so that it steers the behavior of the pretrained ViTs. We visualize the self-attention map at the 3\textsuperscript{rd} (early), 7\textsuperscript{th} (middle), and 12\textsuperscript{th} (late) blocks of SSL ViT with and without our method in Figure~\ref{fig:gate_lt}. MAE with prompt tuning attends to a different region in the image. Especially at the late block, MAE with Gated Prompt Tuning attends to the object's head while MAE attends to the object's boundary. In addition, when we added learnable temperatures, the resulting self-attention was different from when Gated Prompt Tuning was used alone. This suggests that learnable temperature also plays a role in adjusting self-attention.

\begin{figure*}[hbt!]
\centering
\includegraphics[width=1
\linewidth]{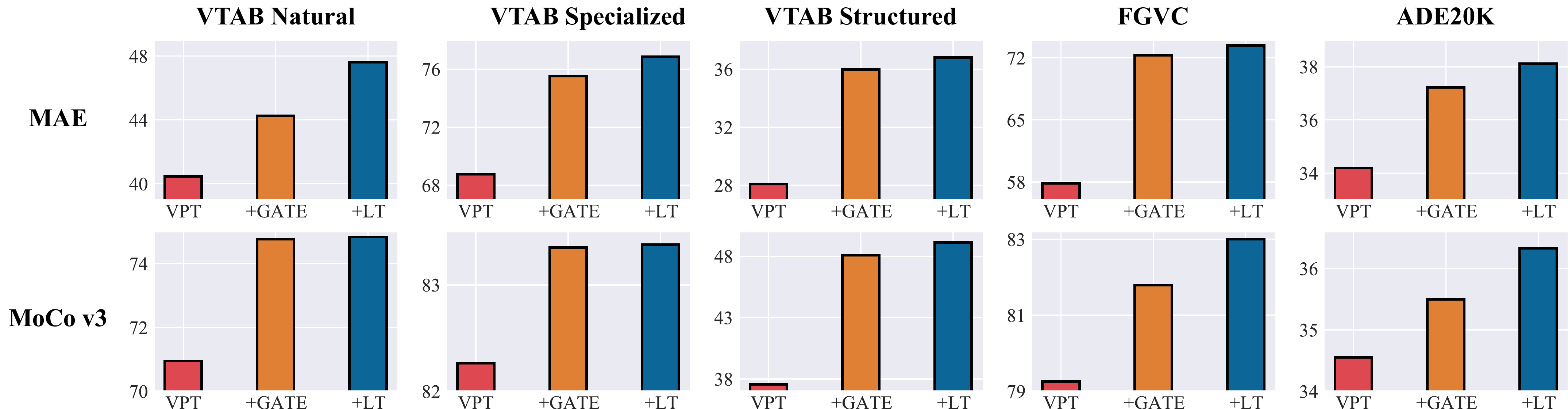}
\vspace*{-0.2cm}
\caption{Ablation study across the benchmarks. For VTAB-1K and FGVC, we report average classification performance, and for ADE20K, we report semantic segmentation performance. GATE denotes Gated Prompt Tuning and LT denotes Adaptive Attention Shaping with learnable temperatures.}
\label{fig:ablation_bar}
\vspace*{0.0cm}
\end{figure*}

\begin{figure}[t]
\centering
\includegraphics[width=1.0
\linewidth]{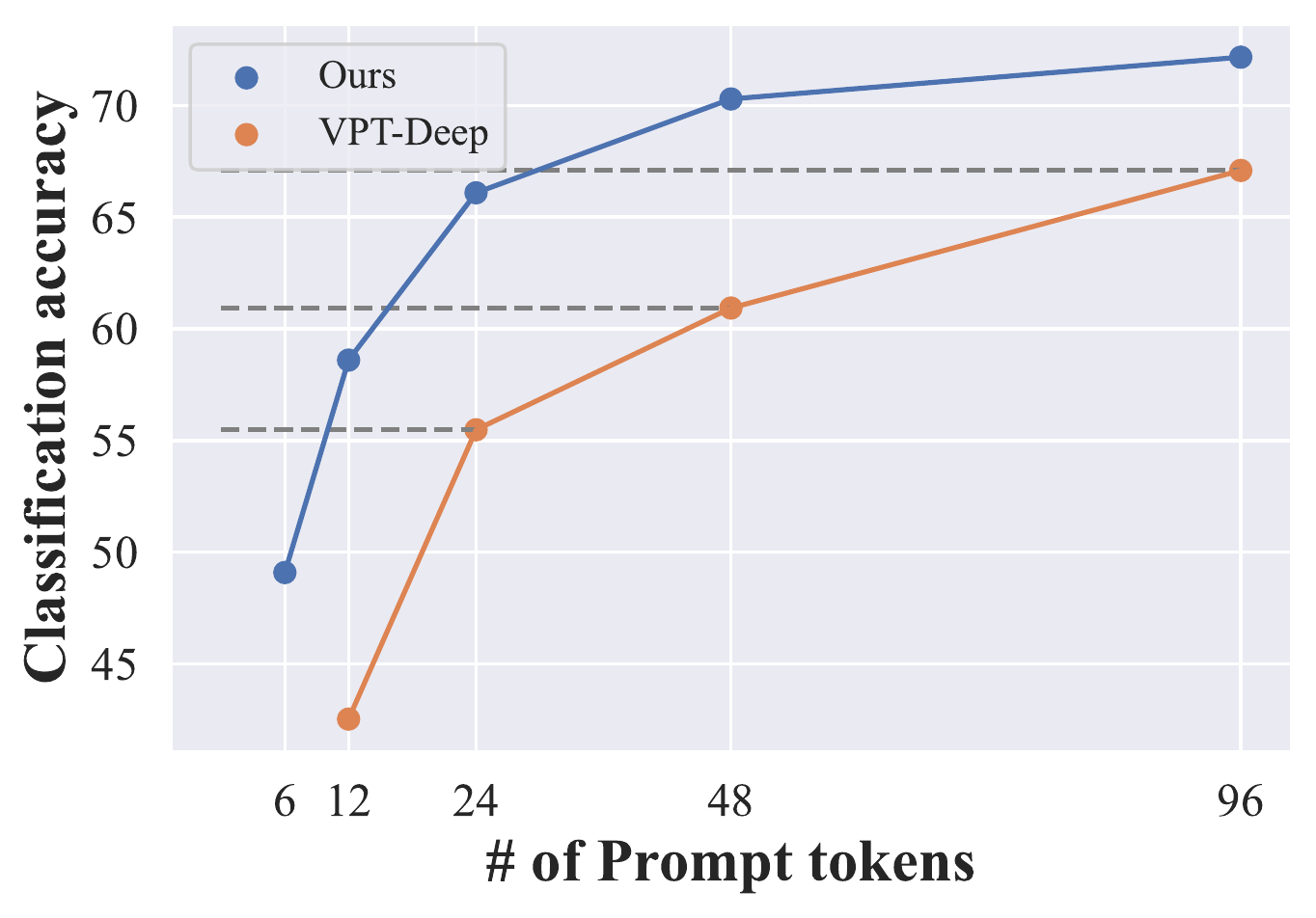}
\vspace*{-1.0em}
\caption{Performance comparison between VPT-deep and our method under the same number of prompt tokens. We used MAE as the SSL ViT backbone and evaluated it on the Stanford Cars dataset. For VPT-deep, using 12, 24, 48, and 96 prompt tokens denotes that 1, 2, 4, and 8 prompt tokens are inserted into each block of ViT-B/16.}
\label{fig:num_token}
\vspace*{0.0cm}
\end{figure}

\subsection{Ablation Studies}
In this section, we conduct ablation studies on the efficacy of our Gated Prompt Tuning and Adaptive Attention Shaping. In Figure~\ref{fig:ablation_bar}, we report the performance changes as our proposed components are added to VPT-shallow. Across all the benchmarks and SSL ViTs, our Gated Prompt Tuning consistently improves performance from VPT-shallow. This verifies that interaction with selective ViT blocks rather than all blocks boosts the strength of prompt tokens for task adaptation. Moreover, Adaptive Attention Shaping with learnable temperatures improves performances in almost all cases. These results support that adjusting self-attention score with adaptive temperature scaling aids prompts to encode improved instructions. The results on each individual dataset are shown in Appendix~\ref{subsec:detail_abl}.

To evaluate the effectiveness of gating operation, we apply the learnable hard gates implemented with Gumbel-Sigmoid~\cite{geng2020does,jang2016categorical} to our proposed method for CUB and OxfordFlowers classification. As shown in Table~\ref{table:gumbel}, using hard gates implemented with Gumbel-Sigmoid outperforms VPT-shallow. This indicates that selective interaction with ViT blocks is effective in task-adaptation using prompt tokens. Our method with soft gates shows improved performances compared to using hard gates. This is because soft gates enable prompt tokens to interact partially with blocks if there is any desirable factor for the task. On the other hand, hard gates would lead to suboptimal results since they exclude the entire block even though it is partially task-relevant.

In addition, we investigate the parameter efficiency of our method using MAE on the Stanford Cars dataset compared to VPT-deep by varying the number of input prompt tokens. In Figure~\ref{fig:num_token}, our method outperforms VPT-deep with only half the number of prompt tokens in all cases. For example, by using only 24 prompt tokens, our method outperforms VPT-deep with 48 prompt tokens (66.1\% vs. 60.9\%). For VPT-deep, it is impossible to use fewer than 12 prompt tokens because it requires at least one prompt token for each ViT block, but our method can handle the number of prompt tokens fewer than 12 and still outperform with fewer prompt tokens. 
In Appendix~\ref{subsec:token_eff}, we provide additional experimental results on the FGVC and ADE20K benchmarks, which demonstrate that our Gated Prompt Tuning employs prompt tokens efficiently for task adaptation.
\begin{table}[t]\centering
\setlength{\tabcolsep}{4pt}
\caption{Ablation study on the gate. We used MAE as Self-supervised ViT. PT denotes prompt token and GH denotes the hard gate implemented with Gumbel-Sigmoid function.}
\label{table:gumbel}
\vskip -0.15in
\begin{center}
\begin{small}
\begin{sc}
\begin{tabular}{lcll}
\toprule
Method &\# of PTs &CUB &Flowers \\
\midrule
VPT-shallow &100 &42.15 &69.15 \\
Ours (w/ GH) &100 &65.46 (\textcolor{blue}{+23.31}) &76.55 (\textcolor{blue}{+7.4})\\
Ours &100 &\textbf{70.56} (\textcolor{blue}{+28.41}) &\textbf{78.55} (\textcolor{blue}{+9.4})\\
\bottomrule
\end{tabular}
\end{sc}
\end{small}
\end{center}
\end{table}

\section{Related Work}\label{sec:relatedwork}
\subsection{Self-supervised Vision Transformers}
Self-supervised Vision Transformers have proven to be an excellent pretrained backbone for computer vision tasks~\cite{bao2021beit, he2022masked, chen2021empirical, xie2022simmim, zhou2021ibot, caron2021emerging}. MoCo v3~\cite{chen2021empirical} and DINO~\cite{caron2021emerging} are representative of the instance-based approach, in which they learn representations that are invariant over random transformations. Masked image modeling~\cite{he2022masked, bao2021beit, zhou2021ibot, xie2022simmim}, which learns representations by recovering randomly masked patches, is another promising branch of self-supervised learning. Utilizing these models for diverse computer vision tasks is a promising strategy, as they have demonstrated excellent transferability and high performance~\cite{he2022masked, chen2021empirical, caron2021emerging, bao2021beit, zhou2021ibot}. However, the tuning methods for self-supervised Vision Transformers during their transfer to downstream vision tasks, particularly prompt-based tuning, have been less explored.

\subsection{Transfer Learning}
Transfer learning aims to efficiently utilize pretrained neural networks for a wide range of downstream tasks. The most basic approach, full fine-tuning, involves training both the pretrained backbone and the task-specific network. Recently, considerable research has been conducted on tuning large pretrained models in a parameter-efficient manner~\cite{jia2022visual, houlsby2019parameter, cai2020tinytl, chen2022adaptformer, bahng2022visual, li2021prefix, lester2021power, huang2023extensibleacl}. 
Cai et al.~\yrcite{cai2020tinytl} proposed to freeze the weights and only update the bias of the pretrained models. Chen et al.~\yrcite{chen2022adaptformer} introduced an additional lightweight module, known as AdaptMLP, in an MLP module of ViT blocks and fine-tune it. Among these, prompt tuning employs learnable perturbations in the embedding space or pixel space~\cite{jia2022visual, bahng2022visual}. However, since these mainly deal with supervised pretrained models, there are few studies on parameter-efficient tuning for self-supervised ViTs. In this work, we develop a prompt-based transfer learning method for self-supervised ViTs.

\subsection{Discussion}
Our proposed method can be interpreted as performing a scaling operation on prompt tokens using a gate. There are previous works that utilize scaling operations, such as AdaptFormer~\cite{chen2022adaptformer}, to achieve efficient fine-tuning of Vision Transformers. However, our work and AdaptFormer diverge in two aspects. First, the location of the scaling operation in AdaptFormer is different from that of our gating operation. In AdaptFormer, the scaling operation is applied to the patch representation, emanating from an additional branch in the MLP module. In contrast, in our proposed method, the gating operation is exclusively conducted for the prompt token and is situated between the Transformer blocks. Second, the purpose of scaling differs. While the scaling operation in AdaptFormer seeks to balance task-agnostic and task-specific features from two distinct branches within each block, our gating operation is designed to regulate the interaction between prompt tokens and each block. To the best of our knowledge, the approach we propose represents one of the first implementations of a gating operation in prompting for computer vision tasks.

In NLP, a concurrent study, known as Prompt Gating~\cite{huang2023extensibleacl}, was recently reported that makes use of trainable gates. This approach aims to combine independently trained prefixes, each learned separately for single-aspect text generation, to enable controllable multiple-aspect text generation during inference. A problem arises in this context where the independently trained prefixes for each aspect interact in an attention sublayer, causing mutual interference and thereby reducing controllability. Prompt Gating addresses this issue by introducing gates that rescale the prefixes for each aspect, adjusting the magnitudes of the prefixes for each aspect within the attention sublayer to alleviate mutual interference. Our approach differs from Prompt Gating in terms of the motivation, objectives, and the implementation of the gating operation. First, our work is driven by the observation that self-supervised ViTs encode more information in the blocks compared to supervised ViTs, thereby making it challenging for prompts to focus on task-relevant blocks. Second, our work aims to facilitate focused interactions between the prompt and the task-relevant blocks, as opposed to adjusting the interaction between multiple task prompts. To accomplish this, our gating operation functions by creating a convex combination of the output prompt representations from the previous block and the current block. In contrast, Prompt Gating merely scales the attention hidden state of the prompt in the current block.

\section{Conclusion}\label{sec:conclusion}
In this work, we propose an enhanced prompt-based transfer method for self-supervised ViTs. The task-relevant blocks in the pretrained ViTs depend on pretraining methods and downstream tasks. To address this, we introduce Gated Prompt Tuning, which adopts learnable gates and directs the prompt to selectively focus on task-relevant blocks for effective task adaptation. Furthermore, we introduce Adaptive Attention Shaping, which adjusts the attention score and further enhances task-specific instruction with prompts. Extensive experimental results across diverse benchmarks confirm that our proposed method more effectively utilizes prompt tokens for task adaptation.
\section*{Acknowledgements}
This work was supported by the National Research Foundation of Korea (NRF) grants funded by the Korea government (Ministry of Science and ICT, MSIT) (2022R1A3B1077720 and 2022R1A5A708390811), SNU-Naver Hyperscale AI Center, Institute of Information \& Communications Technology Planning \& Evaluation (IITP) grants funded by the Korea government (MSIT) (2022-0-00959 and 2021-0-01343: AI Graduate School Program, SNU), and the BK21 FOUR program of the Education and Research Program for Future ICT Pioneers, Seoul National University in 2023.
\bibliography{References}
\bibliographystyle{icml2023}
\newpage
\appendix
\onecolumn

\section{Implementation Details}
\subsection{Pseudo code}
We provide a Pytorch-like pseudo code for our proposed Gated Prompt Tuning in Algorithm~\ref{alg:code}. The gating operation is implemented by adding a few additional lines of code in the block operation of the Vision Transformer.
\begin{algorithm}[h]
\caption{PyTorch-like Pseudocode for Gated Prompt Tuning}
\label{alg:code}
\definecolor{codeblue}{rgb}{0.25,0.5,0.5}
\definecolor{codekw}{rgb}{0.85, 0.18, 0.50}
\lstset{
  backgroundcolor=\color{white},
  basicstyle=\fontsize{10pt}{10pt}\ttfamily\selectfont,
  columns=fullflexible,
  breaklines=true,
  captionpos=b,
  commentstyle=\fontsize{7.5pt}{7.5pt}\color{codeblue},
  keywordstyle=\fontsize{7.5pt}{7.5pt}\color{codekw},
}
\begin{lstlisting}[language=python]
# cls: CLS token
# x: patch tokens
# p: prompt tokens
# gamma: gate priors
# blocks: ViT blocks
# N_p: number of prompt tokens

# prepend prompt tokens
x = cat([cls, p, x], dim=1)
for i, blk in enumerate(blocks): 
    if i == len(blocks) - 1:
        x = blk(x)
    else:
        # compute gate values
        gate = gamma[i].sigmoid() 
        # input prompt representation of i-th block
        prompt_before_block = x[:, 1: 1+N_p, :] 
        
        x = blk(x) 
        # output prompt representation of i-th block
        prompt_after_block = x[:, 1: 1+N_p, :] 
        # gated prompt representation
        gated_prompt = gate * prompt_after_block + (1 - gate) * prompt_before_block
    
        # pass the gated prompt representation to the next block
        x = cat([
             x[:, 0:1, :], 
             gated_prompt, 
             x[:, 1+N_p:, :]
        ], dim=1) 

\end{lstlisting}
\end{algorithm}

\newpage
\subsection{Selected Hyperparameters} \label{detail_hyp}
The hyperparameters used to train the models for FGVC (Table~\ref{table:fgvc}), VTAB-1K~\cite{zhai2019large} (Table~\ref{table:vtab}), and ADE20K (Table~\ref{table:ade}) are listed in Table~\ref{table:hparam}. We used the SGD optimizer, and the learning rate was searched among \{0.05, 0.1, 0.25, 0.5, 1.0, 2.5, 5.0\}. For ADE20K semantic segmentation, we used the default hyperparameters following SETR-PUP~\cite{zheng2021rethinking}.

In Figure~\ref{fig:motivation_acc}, for CUB, we used a learning rate 0.1 for MAE and Supervised ViT, and a learning rate 1.0 for MoCo v3. For KITTI, we used a learning rate 0.1 for all models. In Figure~\ref{fig:num_token}, learning rates used for each model training are the same as those used in Table~\ref{table:fgvc}.
\begin{table*}[h]
  \centering
  \caption{Selected hyper-parameters of our method for each downstream task and SSL method. BS denotes batch size, PT denotes prompt token, LR denotes learning rate and GATE INIT. denotes the initialized gate prior.}
  \vspace{0.2cm}
  \begin{sc}
    \begin{tabular}{l|cccc|cccc}
    \Xhline{1pt}
          & \multicolumn{4}{c|}{MAE}       & \multicolumn{4}{c}{MoCo v3} \\
    
          & BS & \# of PTs & lr & Gate Init. & BS & \# of PTs & lr & Gate Init. \\\hline
    Caltech101 & 128   & 1     & 0.5   & 5     & 128   & 10    & 2.5   & 4 \\
    CIFAR-100 & 128   & 1     & 0.25  & 5     & 128   & 10    & 1     & 15 \\
    Clevr\_Dist & 128   & 1     & 0.1   & 15    & 128   & 5     & 1     & 10 \\
    Clevr\_Count & 128   & 1     & 0.05  & 15    & 128   & 5     & 0.5   & 5 \\
    Retinopathy & 128   & 5     & 0.05  & 10    & 128   & 5     & 0.5   & 5 \\
    DMlab & 128   & 1     & 0.1   & 5     & 128   & 1     & 1     & 5 \\
    dSpr\_Ori & 128   & 1     & 0.05  & 5     & 128   & 5     & 0.5   & 5 \\
    dSpr\_Loc & 128   & 1     & 0.1   & 5     & 128   & 5     & 0.5   & 5 \\
    DTD   & 128   & 1     & 0.25  & 1     & 128   & 10    & 1     & 5 \\
    EuroSAT & 128   & 5     & 0.1   & 1     & 128   & 10    & 1     & 5 \\
    KITTI-Dist & 128   & 5     & 0.1   & 7     & 128   & 5     & 1     & 5 \\
    Flowers102 & 128   & 1     & 0.25  & 5     & 128   & 5     & 1     & 7 \\
    Pets  & 128   & 1     & 0.5   & 10    & 64    & 100   & 1     & 1 \\
    Camelyon & 128   & 1     & 0.05  & 5     & 128   & 1     & 1     & 10 \\
    Resisc45 & 128   & 10    & 0.25  & 5     & 64    & 100   & 1     & 7 \\
    sNORB\_Azim & 128   & 1     & 0.1   & 20    & 128   & 10    & 0.5   & 3 \\
    sNORB\_Elev & 128   & 1     & 0.05  & 10    & 128   & 5     & 0.5   & 4 \\
    Sun397 & 128   & 1     & 0.5   & 10    & 128   & 1     & 1     & 4 \\
    SVHN  & 128   & 1     & 0.1   & 10    & 128   & 5     & 1     & 10 \\
    \hline
    CUB   & 64    & 100   & 0.1   & 5     & 64    & 100   & 1     & 5 \\
    Flowers & 64    & 100   & 0.1   & 15    & 64    & 100   & 1     & 10 \\
    Cars  & 64    & 100   & 0.25  & 5    & 64    & 100   & 0.5   & 5 \\
    Dogs  & 64    & 100   & 0.5   & 15    & 64    & 100   & 0.5   & 10 \\
    NABirds & 64    & 100   & 0.5   & 5     & 64    & 100   & 1     & 5 \\
    \hline
    ADE20K & 16    & 100   & 0.001 & 10     & 16    & 100   & 0.001 & 10 \\
    \Xhline{1pt}
    \end{tabular}%
    \end{sc}
  \label{table:hparam}%
\end{table*}%

\newpage
\section{Additional Experiments}
\subsection{Experiments with Fewer Prompt Tokens}
\label{subsec:token_eff}
To demonstrate that our Gated Prompt Tuning employs prompt tokens efficiently for task adaptation, we conducted additional experiments on FGVC image classification and ADE20K semantic segmentation. For ADE20K semantic segmentation we used 24 prompt tokens. In Table~\ref{table:ade-token}, our proposed method still outperforms VPT-deep when using fewer tokens.

\begin{table*}[h]\centering
\caption{Semantic segmentation results on ADE20K with fewer tokens. PT denotes prompt token.}
\label{table:ade-token}
\vskip 0.15in
\begin{center}
\begin{small}
\begin{sc}
\begin{tabular}{llccc}
\toprule
\multirow{2}{*}{SSL} &\multirow{2}{*}{Method} & \multirow{2}{*}{\# of PTs} &\multicolumn{1}{p{1cm}}{\centering mIou \\ (SS)} &\multicolumn{1}{p{1cm}}{\centering mIoU \\ (ms)} \\
\midrule
\multirow{3}{*}{MAE} 
&VPT-shallow &24 &34.77 &35.93 \\
&VPT-deep &24 &37.03 &38.25 \\
&Ours  &24 &\textbf{38.13} &\textbf{39.23} \\
\midrule
\multirow{3}{*}{MoCo v3} 
&VPT-shallow &24 &34.75 &36.34 \\
&VPT-deep &24  &35.96 &37.47 \\
&Ours  &24 &\textbf{37.02} &\textbf{38.57} \\
\bottomrule
\end{tabular}
\end{sc}
\end{small}
\end{center}
\end{table*}
For the FGVC classification, we evaluated the performance of our proposed method and VPT-deep using 24, 48, and 96 prompt tokens. We utilized MAE as the self-supervised ViT in this experiment. We observed that the average performance gap between our method and VPT-deep widens when using fewer tokens in Table~\ref{table:fgvc-token}. The hyperparameters used for each model training are the same as those used in Table~\ref{table:fgvc}, except for the number of prompt tokens.
\begin{table*}[h]
\caption{Classification results on FGVC with fewer tokens. PT denotes prompt token.} 
\label{table:fgvc-token}
\vskip 0.15in
\begin{center}
\begin{small}
\begin{sc}
\begin{tabular}{clcccccl}
\toprule
\# of PTs  &Method  & CUB   & Flowers & Cars  & Dogs  & NABirds & AVG\\
\midrule
\multirow{2}{*}{24}  
&VPT-deep  &59.68 &67.51 &57.68 &\textbf{77.91} &54.65 &63.49 \\
&Ours &\textbf{65.26} &\textbf{73.80} &\textbf{66.09} &75.72 &\textbf{62.10} &\textbf{68.59} (\textcolor{blue}{+5.10})\\
\midrule
\multirow{2}{*}{48}  
&VPT-deep  &66.43 &71.31 &64.06 &\textbf{78.89} &60.62 &68.26 \\
&Ours  &\textbf{69.00} &\textbf{74.71} &\textbf{70.29} &76.77 &\textbf{63.73} &\textbf{70.90} (\textcolor{blue}{+2.64}) \\
\midrule
\multirow{2}{*}{96}
&VPT-deep  &70.85 &\textbf{76.29} &67.63 &\textbf{78.38} &62.99 &71.23 \\
&Ours  &\textbf{71.40} &75.96 &\textbf{72.16} &77.81 &\textbf{66.25} &\textbf{72.72} (\textcolor{blue}{+1.49}) \\
                  
\bottomrule
\end{tabular}
\end{sc}
\end{small}
\end{center}
\vskip -0.1in
\end{table*}

\subsection{Results on Alternative Vision Transformer Backbones}

In this section, we provide experimental results on other Vision Transformer variants other than ViT-B. Due to the lack of public pretrained models, we used MoCo v3 for ViT-S and MAE for ViT-L. We performed experiments on CUB and OxrfordFlowers classification. The pretrained model checkpoints are obtained from the official repositories of MoCo v3 and MAE~\cite{chen2021empirical, he2022masked}.

In Table~\ref{table:mae-vit-l} and Table~\ref{table:moco-v3-vit-s}, our method outperforms VPT in both CUB and OxfordFlowers datasets in all the ViT variants. In particular, for ViT-L (MAE), there was a significant performance gap between our method and VPT-deep.
\begin{table*}[h]
    \caption{CUB and OxfordFlowers classification results on ViT variants. PT denotes prompt token.}
    \vskip 0.15in
    \begin{subtable}[h]{0.5\textwidth}
    \begin{center}
    \begin{small}
    \begin{sc}
    \caption{ViT-L (MAE)} 
    \begin{tabular}{lccc}
    \toprule
     &\#of PTs &CUB &Flowers\\
    \midrule
    VPT-shallow & 48 & 39.26 &62.77 \\
    VPT-deep & 48 &70.54 &65.44 \\
    Ours & 48 &\textbf{72.99} &\textbf{74.71} \\
    \bottomrule
    \end{tabular}
    \label{table:mae-vit-l}
    \end{sc}
    \end{small}
    \end{center}
    \end{subtable}
    \hfill
    \begin{subtable}[h]{0.5\textwidth}
    \begin{center}
    \begin{small}
    \begin{sc}
    \caption{ViT-S (MoCo v3)} 
    \begin{tabular}{lccc}
    \toprule
     &\#of PTs &CUB &Flowers\\
    \midrule
    VPT-shallow & 48 &68.62 &84.65 \\
    VPT-deep & 48 &73.18 &89.53 \\
    Ours & 48 &\textbf{73.3} &\textbf{91.14} \\
    \bottomrule
    \end{tabular}
    \label{table:moco-v3-vit-s}
    \end{sc}
    \end{small}
    \end{center}
    \end{subtable}
\end{table*}

\newpage
\newpage
\subsection{Detailed Ablation Studies}
\begin{figure}[hbt!]
\centering
\vspace{-0.2cm}

\begin{subfigure}[hbt!]{0.95\textwidth}
    \includegraphics[width=\textwidth]{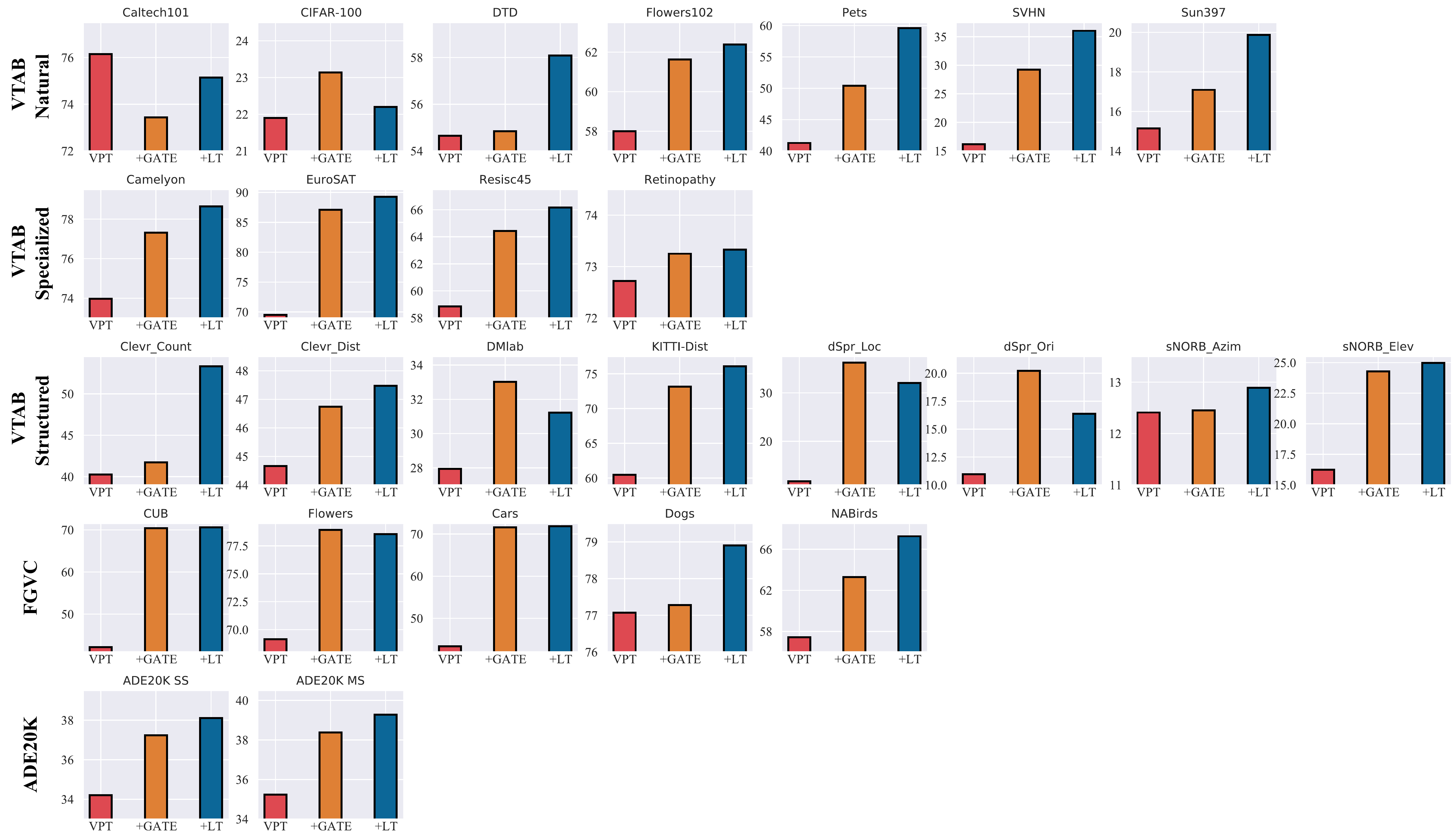}
    \vspace*{-0.8cm}
    \caption{MAE}
    \vspace*{0.2cm}
    \label{fig:abl_mae}
\end{subfigure}
\begin{subfigure}[hbt!]{0.95\textwidth}
    \includegraphics[width=\textwidth]{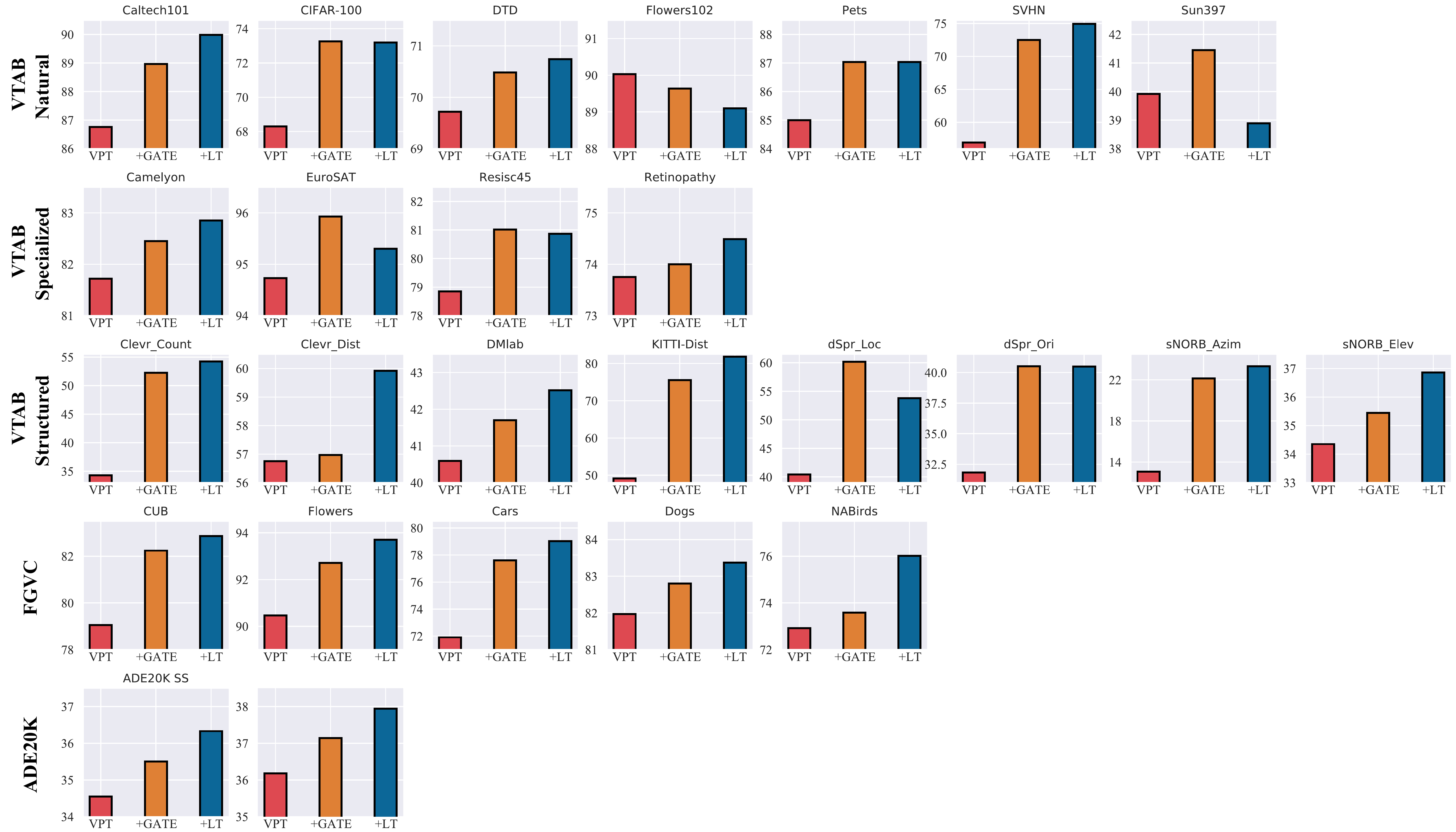}
    \vspace*{-0.6cm}
    \label{fig:abl_moco}
    \caption{MoCo v3}
\end{subfigure}
\caption{Ablation study across the benchmarks. GATE denotes Gated Prompt Tuning and LT denotes Adaptive Attention Shaping with learnable temperatures.}
\label{fig:abl_full}
\vspace*{-0.4cm}
\end{figure}

In this section, we provide ablation results of our proposed method on each individual dataset in FGVC, VTAB-1K, and ADE20K in Figure~\ref{fig:abl_full}.
\label{subsec:detail_abl}

\newpage
\section{Empricial Observations}
\begin{figure}[h]
\centering
\begin{subfigure}[h]{0.48\textwidth}
    \includegraphics[width=\textwidth]{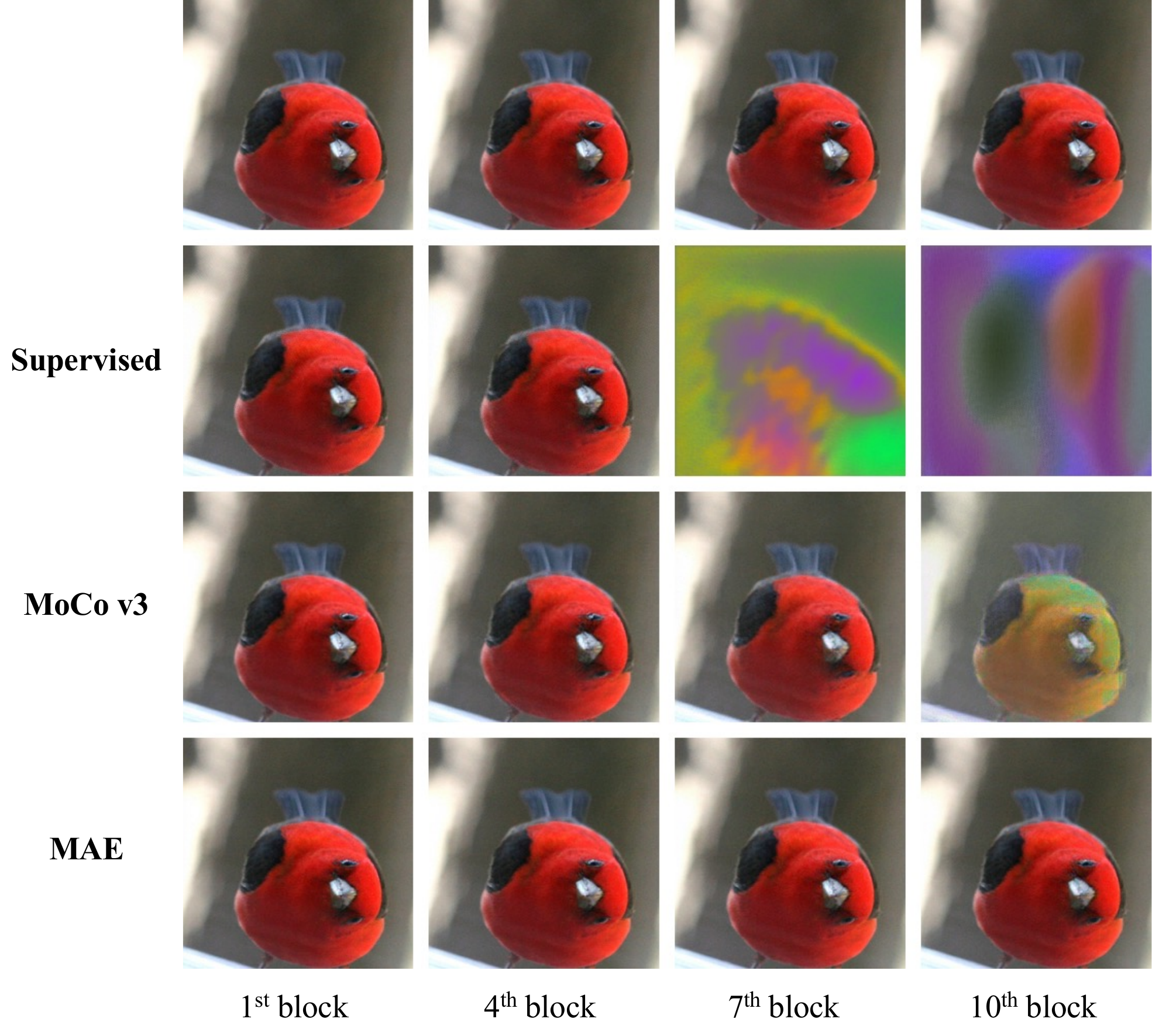}
    \vspace*{0.2cm}
    \label{fig:dip-cub-1}
\end{subfigure}
\hfill
\begin{subfigure}[h]{0.48\textwidth}
    \includegraphics[width=\textwidth]{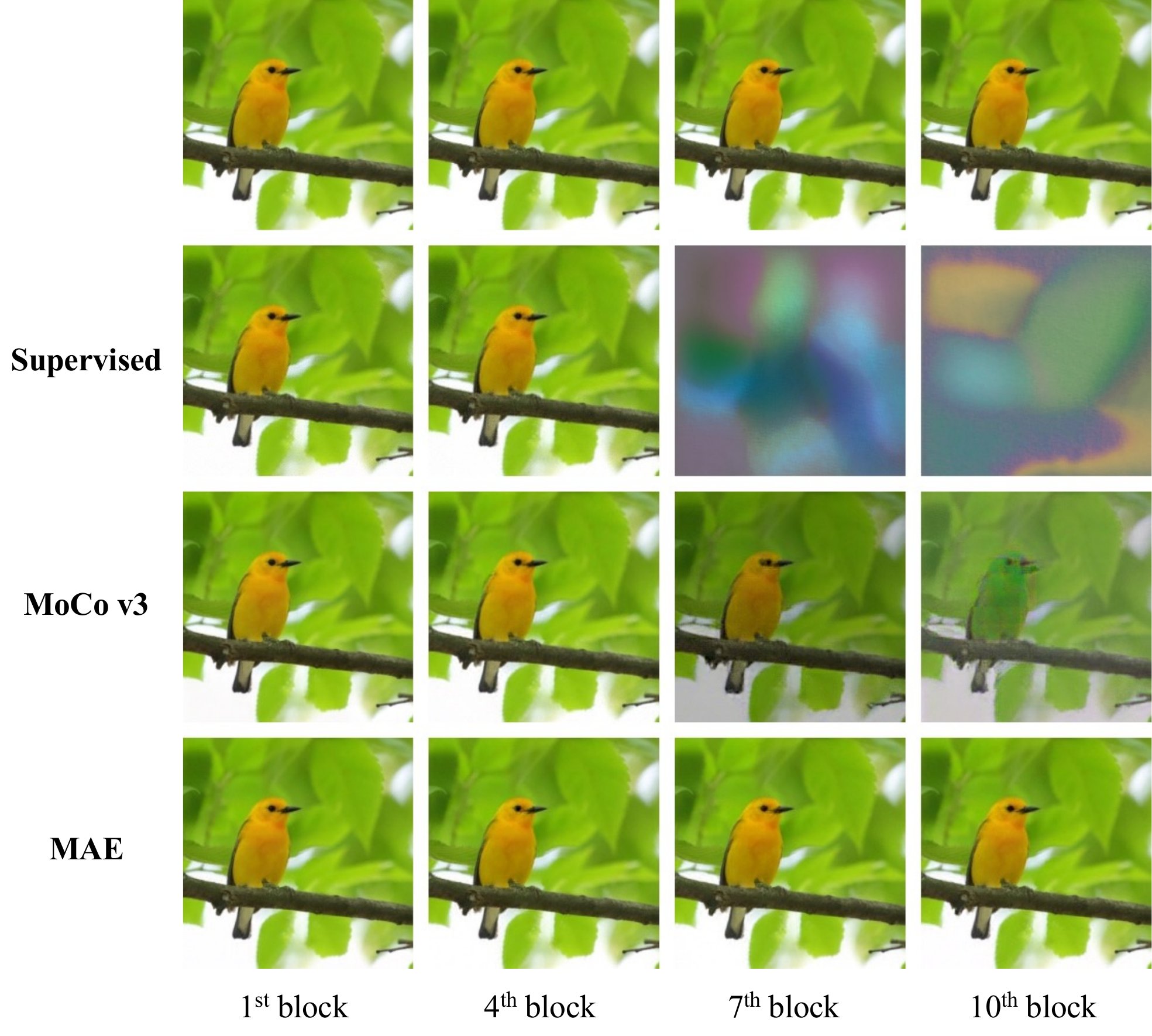}
    \vspace*{0.2cm}
    \label{fig:dip-cub-2}
\end{subfigure}
\begin{subfigure}[h]{0.48\textwidth}
    \includegraphics[width=\textwidth]{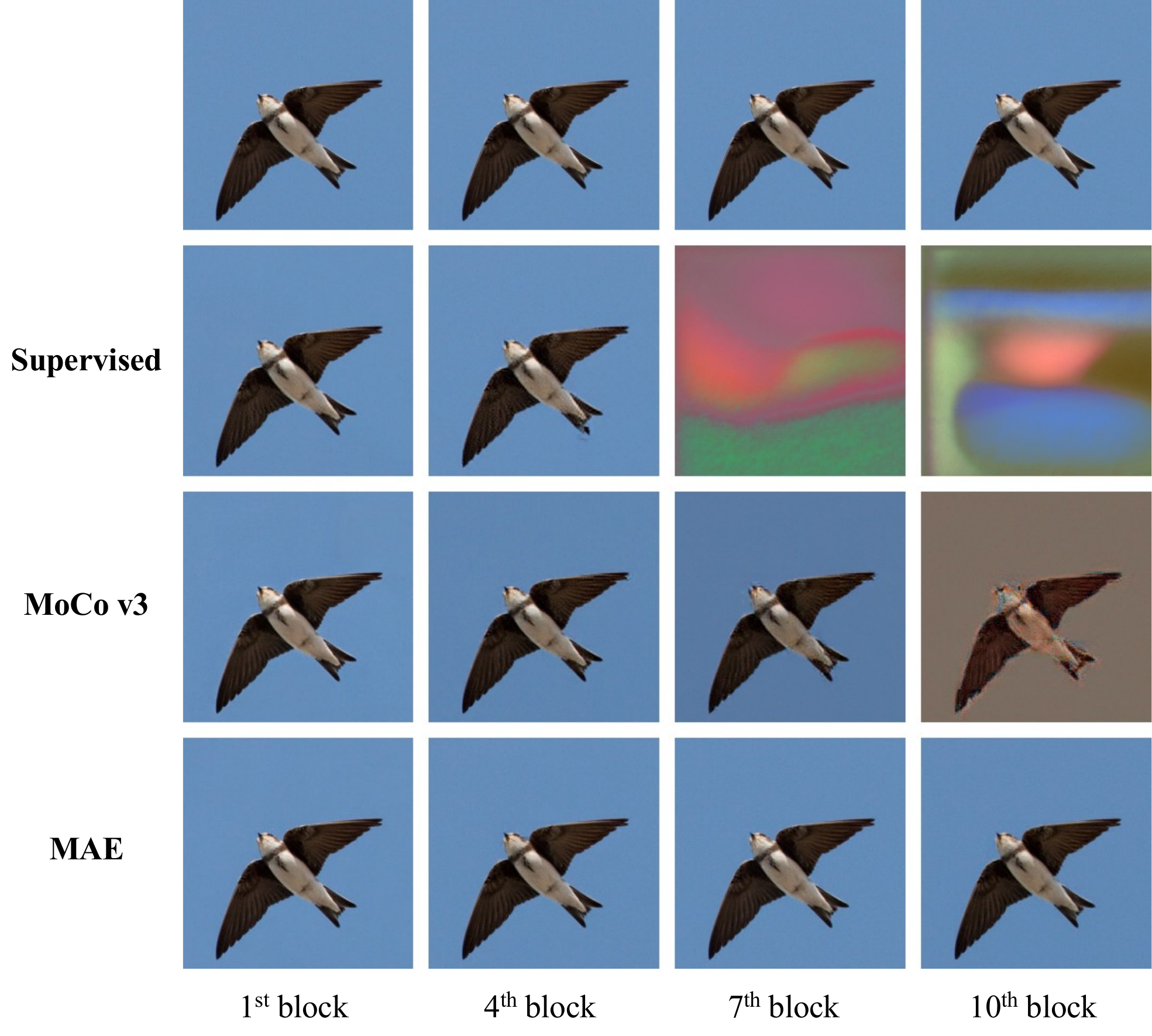}
    \label{fig:dip-cub-3}
\end{subfigure}
\hfill
\begin{subfigure}[h]{0.48\textwidth}
    \includegraphics[width=\textwidth]{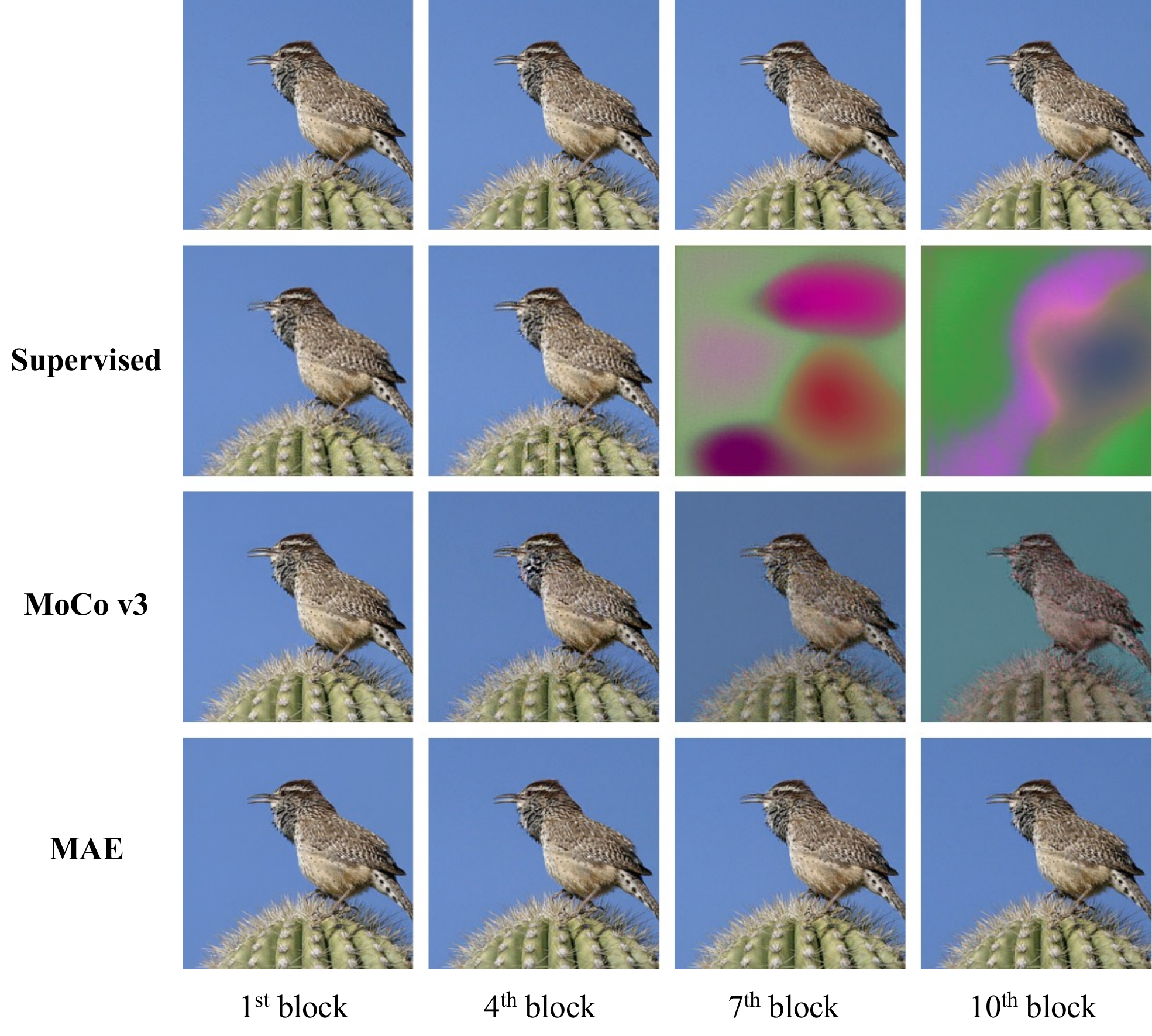}
    \label{fig:dip-cub-4}   
\end{subfigure}
\caption{Reconstructed images using Deep Image Prior (DIP) with pretrained ViT block's representation as a training target. \textbf{Row 1}: original image. \textbf{Rows 2-4}: reconstruction results for each pretrained ViTs.}
\label{fig:dip-cub}
\vspace*{-1.0cm}
\end{figure}

\begin{figure}[h]
\centering
\begin{subfigure}[h]{0.48\textwidth}
    \includegraphics[width=\textwidth]{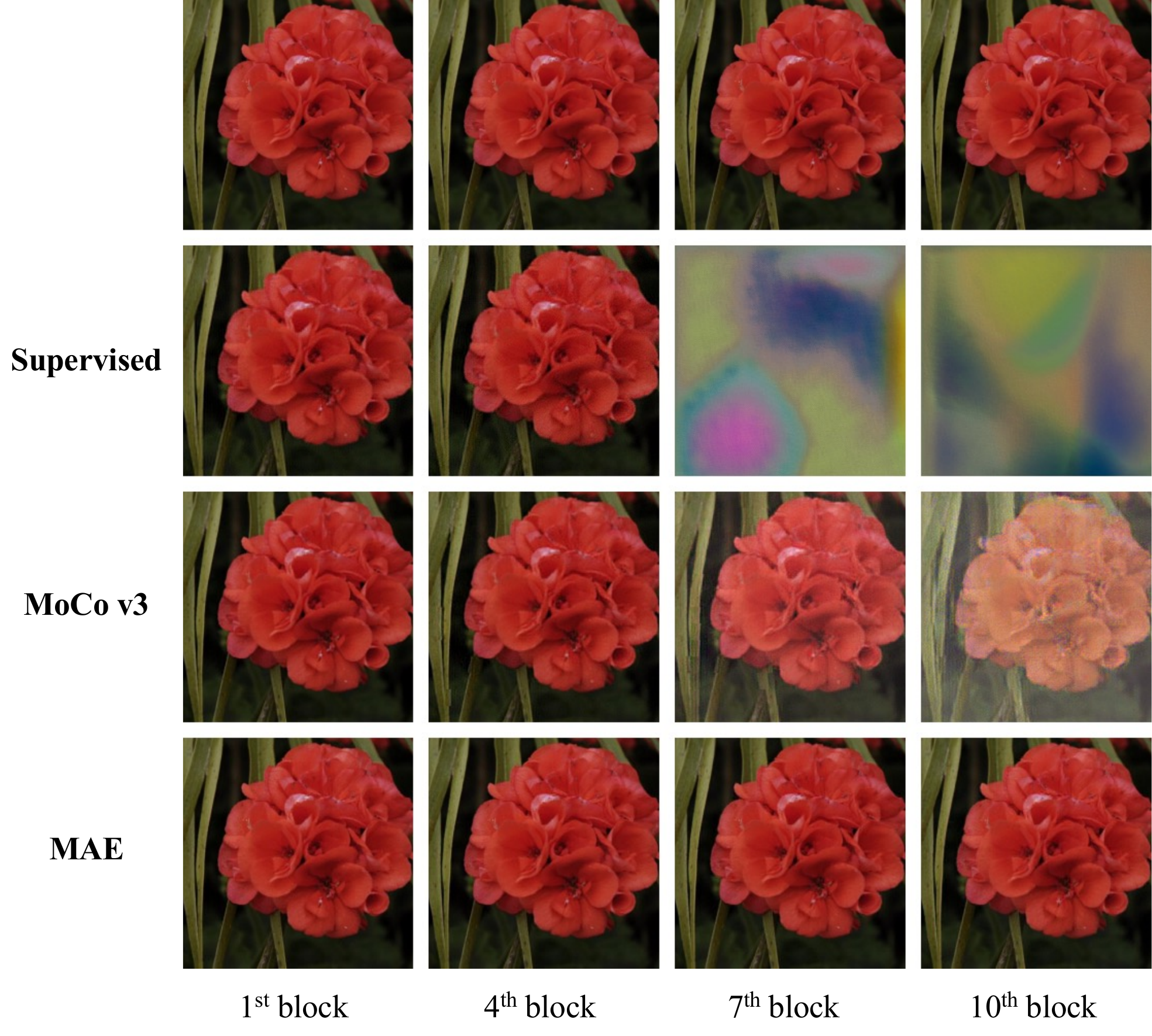}
    \vspace*{0.2cm}
    \label{fig:dip-flowers-1}
\end{subfigure}
\hfill
\begin{subfigure}[h]{0.48\textwidth}
    \includegraphics[width=\textwidth]{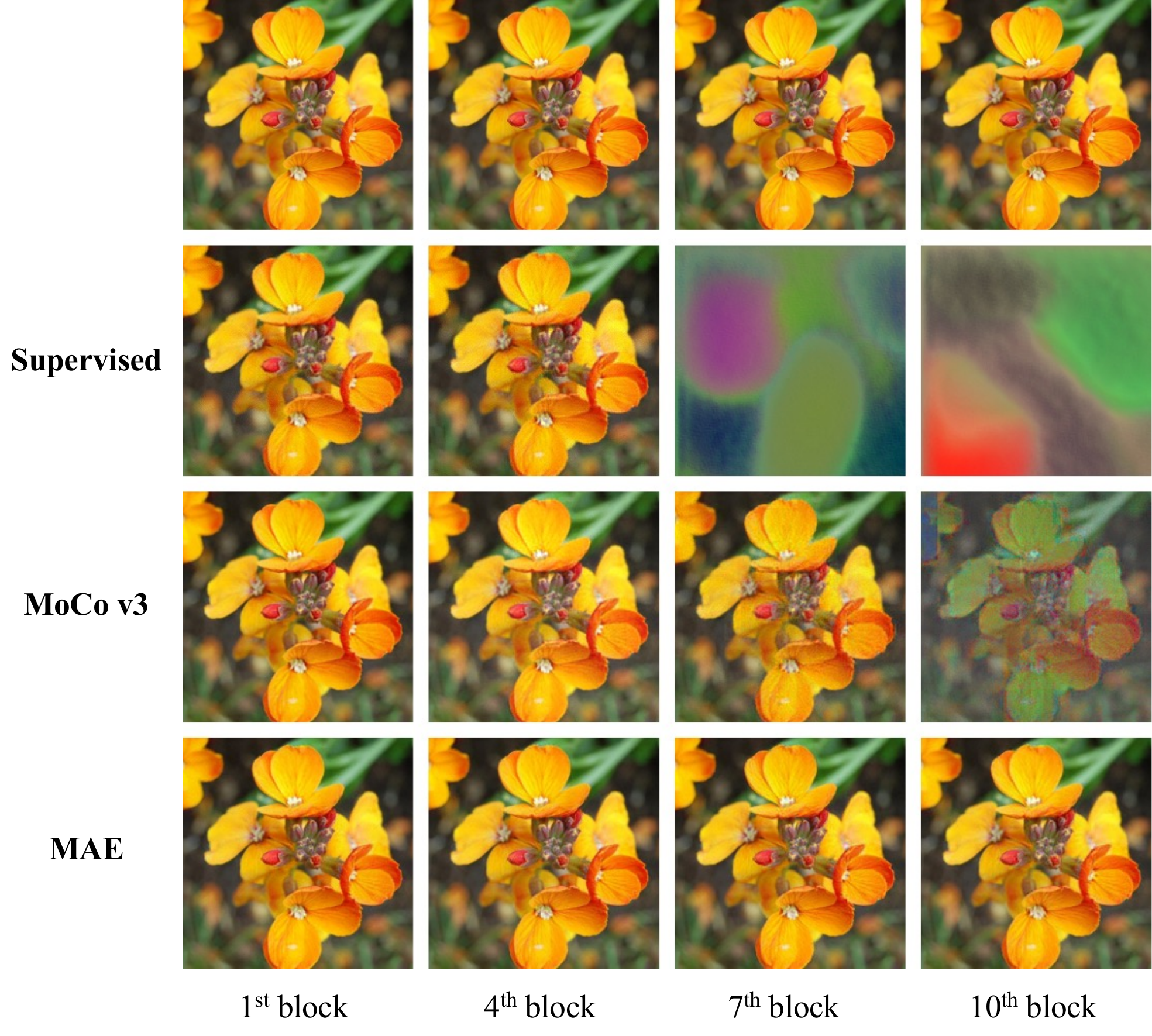}
    \vspace*{0.2cm}
    \label{fig:dip-flowers-2}
\end{subfigure}
\begin{subfigure}[h]{0.48\textwidth}
    \includegraphics[width=\textwidth]{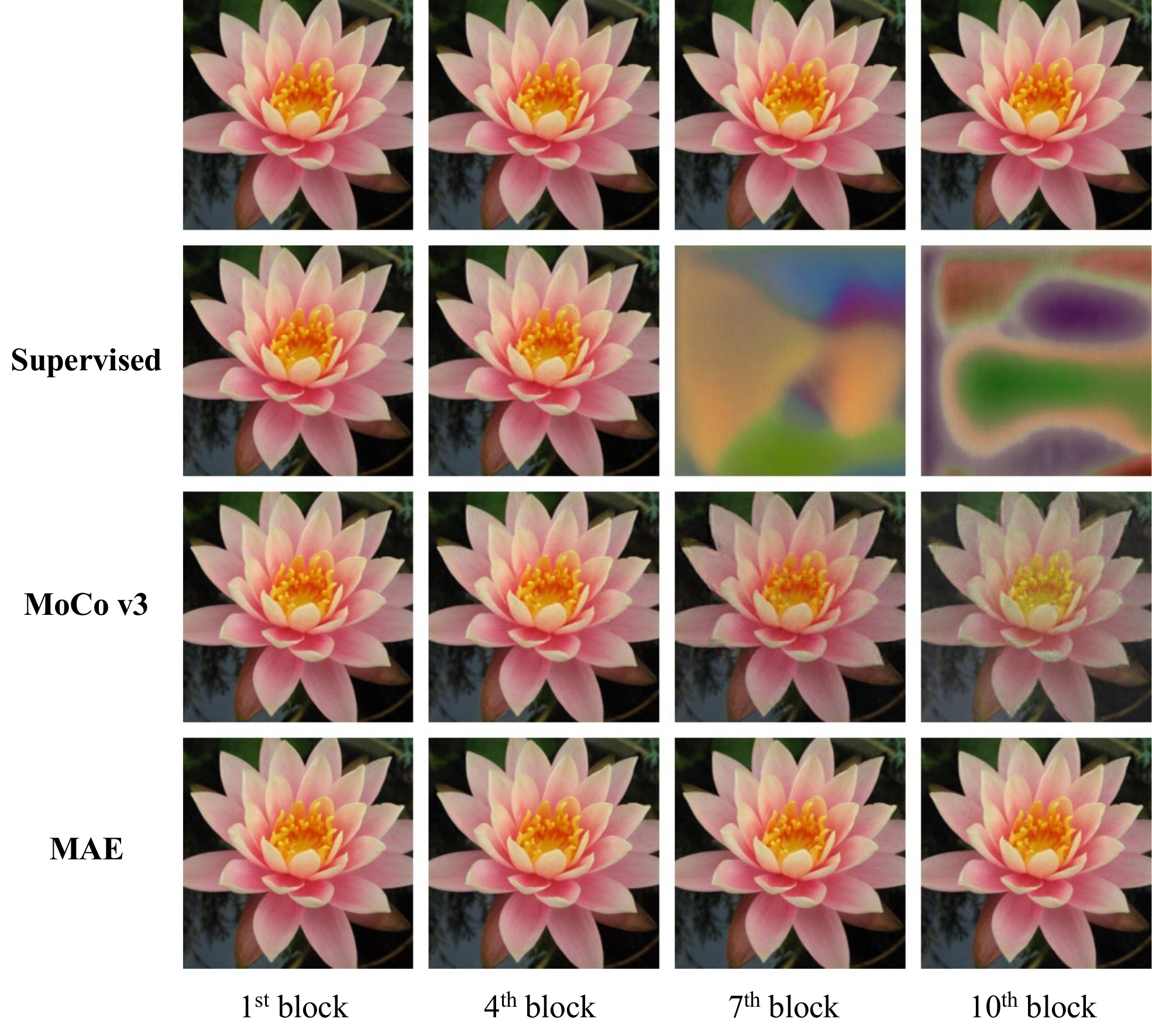}
    \label{fig:dip-flowers-3}
\end{subfigure}
\hfill
\begin{subfigure}[h]{0.48\textwidth}
    \includegraphics[width=\textwidth]{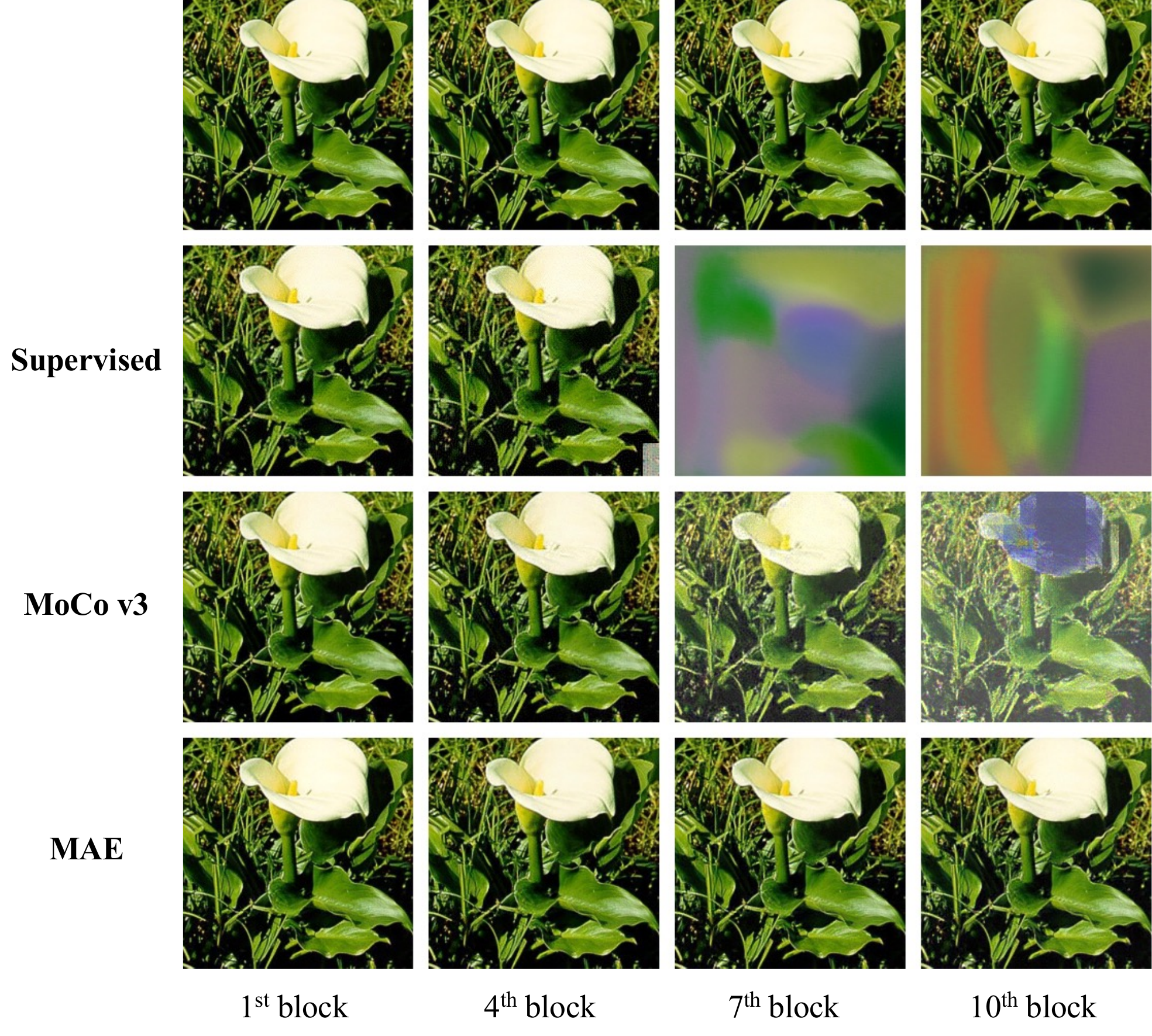}
    \label{fig:dip-flowers-4}   
\end{subfigure}
\caption{Reconstructed images using Deep Image Prior (DIP) with pretrained ViT block's representation as a training target. \textbf{Row 1}: original image. \textbf{Rows 2-4}: reconstruction results for each pretrained ViTs.}
\label{fig:dip-flowers}
\vspace*{0.1cm}
\end{figure}

In this section, we present more qualitative and quantitative experimental results of Deep Image Prior (DIP)~\cite{ulyanov2018deep} applied on the representations of each block in the pretrained ViTs. For supervised ViT, we used pretrained model checkpoint from the official repository of Visual Prompt Tuning~\cite{jia2022visual}. In Figure~\ref{fig:dip-cub} and Figure~\ref{fig:dip-flowers}, it can be observed that the DIP results of each block differ for supervised ViT, MoCo v3, and MAE for both the CUB and OxfordFlowers datasets.
\subsection{PSNR and SSIM scores}
\begin{figure}[h]
\centering

\begin{subfigure}[h]{0.8\textwidth}
    \includegraphics[width=\textwidth]{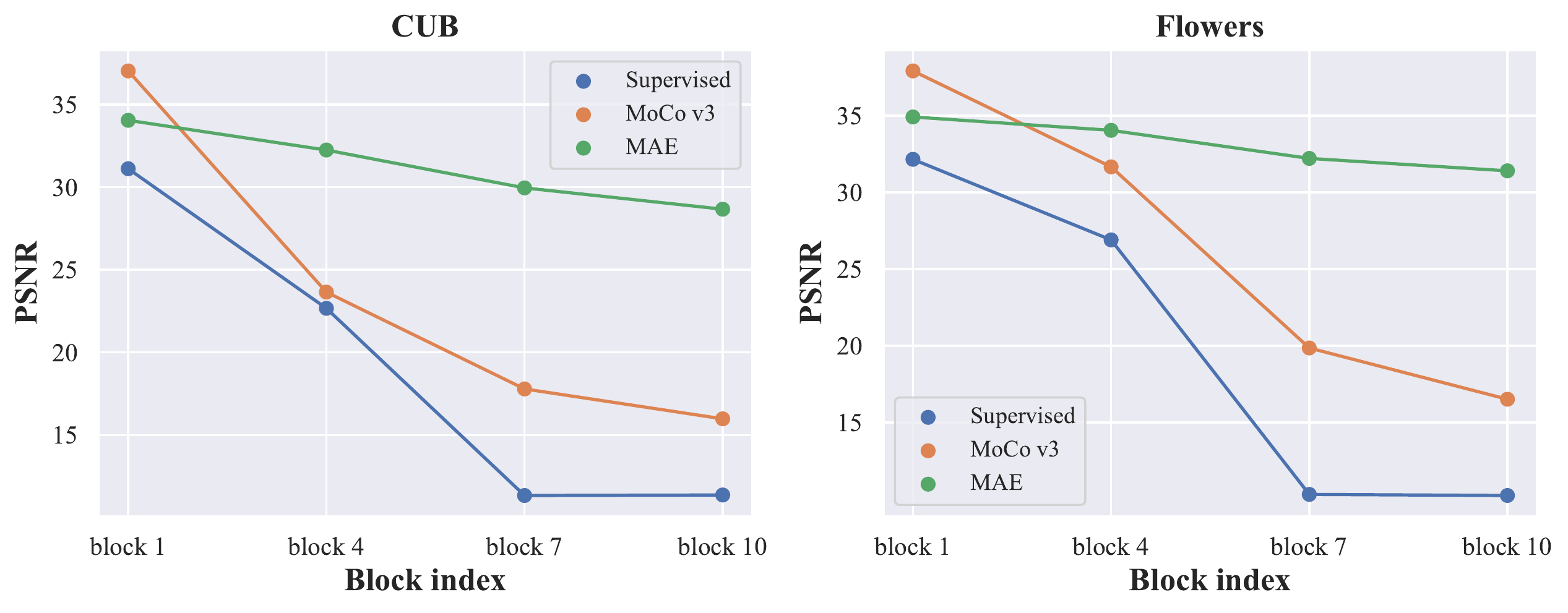}
    \vspace*{-0.6cm}
    \caption{PSNR}
    \vspace*{0.4cm}
\end{subfigure}
\begin{subfigure}{0.8\textwidth}
    \includegraphics[width=\textwidth]{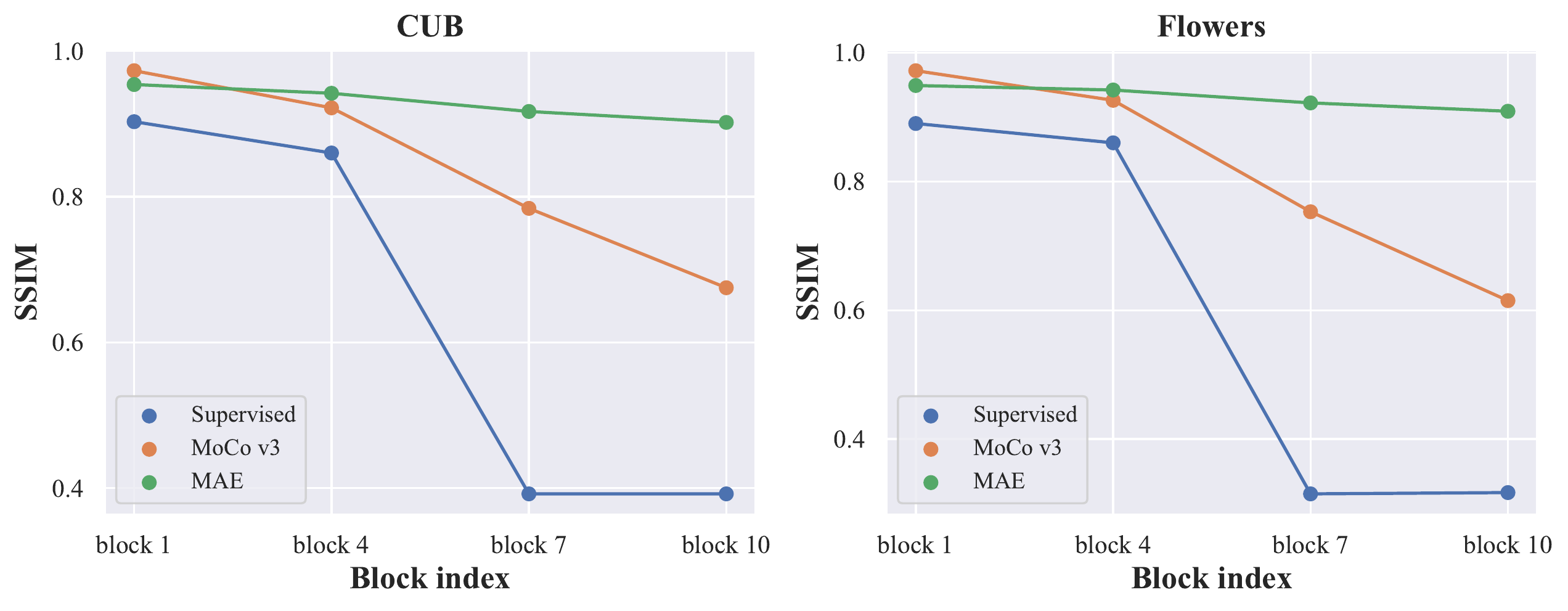}
    \vspace*{-0.6cm}
    \caption{SSIM}
    \vspace*{0.2cm}
\end{subfigure}
\caption{PSNR and SSIM scores for the reconstructed images using Deep Image Prior.}
\label{fig:ps_score}
\vspace*{0.1cm}
\end{figure}

As a quantitative result, we measured the PSNR and SSIM scores of 100 images generated by DIP for each block of the pretrained ViTs. As shown in Figure~\ref{fig:ps_score}, it can be inferred that self-supervised ViTs retain relatively more information, even in the later blocks, resulting in higher-quality reconstructed images.
\label{subsec:emp_quan}

\newpage
\subsection{Deep Image Prior (DIP) results}
\label{subsec:dip_results}

\newpage

\end{document}